\pdfoutput=1
\documentclass{article}

\PassOptionsToPackage{numbers, compress}{natbib}
\usepackage[main, final]{neurips_2026}
\usepackage{bm}
\usepackage[utf8]{inputenc}
\usepackage[T1]{fontenc}
\usepackage[hidelinks]{hyperref}
\usepackage{fontawesome5}
\usepackage{url}
\usepackage{booktabs}
\usepackage{amsfonts}
\usepackage{nicefrac}
\usepackage{microtype}
\usepackage{xcolor}
\usepackage{mathtools}
\usepackage{graphicx}
\usepackage{multirow}
\usepackage{rotating}
\usepackage{pifont}
\usepackage{float}
\usepackage{capt-of}
\usepackage{adjustbox}
\newcommand{\cmark}{\textcolor{green!60!black}{\ding{51}}}
\newcommand{\xmark}{\textcolor{red!70!black}{\ding{55}}}

\hypersetup{
  pdftitle={CRePE: Curved Ray Expectation Positional Encoding for Unified-Camera-Controlled Video Generation},
  pdfauthor={Seonghyun Jin, Youngmin Kim, Sunwoo Park, Jong Chul Ye}
}

\definecolor{crepelink}{HTML}{3D5A80}
\title{CRePE: Curved Ray Expectation Positional Encoding for Unified-Camera-Controlled Video Generation}

\author{%
  Seonghyun Jin$^{1,*}$\qquad
  Youngmin Kim$^{1,2,*,\ddagger}$\qquad
  Sunwoo Park$^{1,*}$\qquad
  Jong Chul Ye$^{1,\dagger}$\\
  $^{1}$Kim Jaechul Graduate School of Artificial Intelligence, KAIST, Seoul, Republic of Korea\\
  $^{2}$Korea University, Seoul, Republic of Korea\\
  \texttt{\{jinotter3,sunwoo\_p,jong.ye\}@kaist.ac.kr}\quad
  \texttt{zeromin03@korea.ac.kr}\\[6pt]
  \href{https://crepe-neurips2026.github.io/}{\textcolor{crepelink}{\faGlobe\ Project Page}}\qquad
  \href{https://github.com/jinotter3/CRePE}{\textcolor{crepelink}{\faGithub\ Code}}
}

\begin{document}
\maketitle

\begingroup
\renewcommand{\thefootnote}{\fnsymbol{footnote}}
\footnotetext[1]{Equal contribution.}
\footnotetext[2]{Corresponding author.}
\footnotetext[3]{Work done during an internship at KAIST.}
\endgroup

\begin{figure}[h]
\centering
\includegraphics[width=0.9\linewidth]{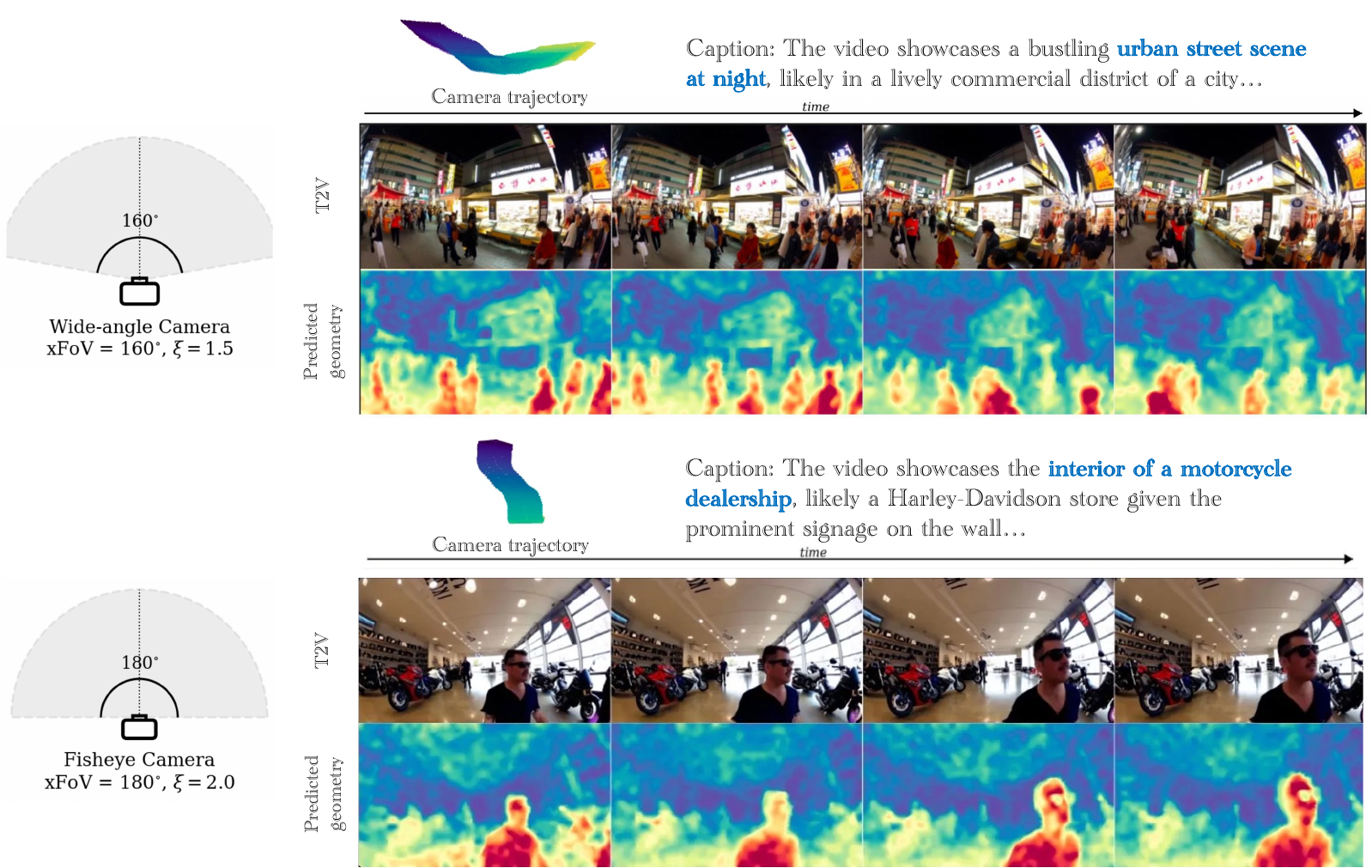}
\caption{\textbf{Fisheye and wide-lens camera-conditioned video generation.} Each row shows a video generated by \textbf{CRePE} under a non-pinhole camera trajectory, alongside the internal geometry map the model estimates and utilizes during generation --- the per-token expected radial distance along its viewing ray, integrated into the rotary phasor at each denoising step. Top: wide-angle camera trajectory with the corresponding generated frames and geometry map. Bottom: fisheye trajectory with its generated frames and geometry map. The geometry map is internal to CRePE rather than an external input: it is a latent positional variable decoded from the geometry head, not the output of a depth decoder or a metric depth estimate.}
\label{fig:main}
\end{figure}

\begin{abstract}
Camera-conditioned video generation requires positional encoding that remains reliable under changes in camera motion, lens configuration, and scene structure.
However, existing attention-level camera encodings either provide ray-only camera signals or rely on pinhole camera geometry, limiting their applicability to general camera control under the Unified Camera Model, including wide-angle and fisheye lenses.
To address this limitation, we propose \textbf{Curved Ray Expectation Positional Encoding (CRePE)}.
CRePE represents each image token as a depth-aware positional distribution along its source ray, providing a Unified Camera Model-compatible positional encoding that naturally captures the curved-ray geometry of wide-angle and fisheye cameras.
CRePE is implemented through a \textbf{Geometric Attention Adapter} added to frozen video DiTs, injecting token-wise scene-distance information into selected attention layers and stabilizing it with pseudo supervision from a monocular geometry foundation model.
This design leads to more stable camera control with competitive video quality in camera-conditioned video generation under the Unified Camera Model, and the resulting model transfers zero-shot to real fisheye and diverse pinhole videos.
Controlled positional-encoding ablations show improvements over a RayRoPE-style endpoint approximation, demonstrating the effectiveness of curved-ray-aware positional encoding, particularly for non-pinhole cameras.
Furthermore, by extending the same positional-encoding pathway to external geometry control through \textbf{Radial MixForcing}, CRePE supports external radial-map control for scene-geometry-conditioned generation and source-video motion transfer beyond camera control.
\end{abstract}

\section{Introduction}
\label{sec:introduction}

Camera-conditioned video generation~\cite{bai2025recammaster} requires models to synthesize videos that follow specified camera trajectories and lens configurations.
Recent studies have expanded the scope of camera conditioning from direct camera conditioning to attention-level positional encoding~\cite{wu2026rayropeprojectiveraypositional,zhang2025unified}.
By incorporating camera geometry directly into positional encoding, video diffusion transformers~\cite{peebles2023scalablediffusionmodelstransformers,wan2025wan} can more explicitly model the relative positional relationships among latent tokens under view changes.
In particular, UCPE~\cite{zhang2025unified} encodes the viewing ray of each token based on a unified central-camera representation, demonstrating the possibility of camera control under the Unified Camera Model (UCM), including wide-angle and fisheye cameras.
This marks an important step toward handling diverse lens models within a unified positional-encoding framework.

However, relative-ray-based positional encoding still has a limitation.
A viewing ray indicates the direction associated with a token, but it does not specify where the underlying scene content lies along that ray.
When the camera moves, token correspondences depend not only on ray direction but also on the scene position along the ray.
Thus, while ray-only encoding provides a stable way to inject camera geometry, it remains limited in representing more refined positional relationships coupled with scene geometry.
This limitation becomes more pronounced for wide-angle and fisheye cameras, where projection geometry changes nonlinearly.

To address this problem, we propose \textbf{Curved Ray Expectation Positional Encoding (CRePE)}.
Instead of representing each token only by its viewing ray, CRePE models it as a depth-aware positional distribution along the source ray, reflecting possible scene positions associated with the token.
CRePE is designed to be compatible with the Unified Camera Model, naturally capturing the nonlinear projection geometry that arises under diverse lens configurations.
This allows the positional encoding to account for how tokens may align with the underlying scene structure and to refine token relationships under view changes.
CRePE builds on RayRoPE's principle of averaging RoPE over ray-position uncertainty~\cite{wu2026rayropeprojectiveraypositional}, replacing its pinhole endpoint interval with a log-radial distribution projected through the UCM (Table~\ref{tab:method_comparison}).

To apply CRePE to camera-conditioned video generation, we introduce a \textbf{Geometric Attention Adapter} for frozen video DiTs.
The adapter predicts token-wise scene distance and incorporates it into the positional relationships used in attention.
Pseudo radial-distance supervision from a monocular geometry foundation model serves as an auxiliary geometric anchor, helping the predicted geometry signal remain stable during training; it is not required at inference.
Through this design, CRePE extends ray-only positional encoding into a scene-geometry-aware positional encoding while preserving the adapter-based camera-conditioning structure.

We evaluate CRePE against camera-conditioned video generation baselines~\cite{bai2025recammaster, zhang2025unified} and analyze the effect of the positional encoding itself through controlled ablations.
CRePE improves overall camera control under the Unified Camera Model with competitive video quality, while showing a generally favorable performance trend over a RayRoPE-style endpoint approximation in the controlled setting.
Furthermore, inspired by YoNoSplat~\cite{ye2026yonosplat}, we extend the same positional-encoding pathway to external geometry control through \textbf{Radial MixForcing}; CRePE aligns tokens with geometry-aware positional information in a compatible form, suggesting its potential for external radial-map control in scene-geometry-conditioned generation and source-video motion transfer. Quantitative DAVIS and non-pinhole PanShot experiments verify adherence to the supplied maps.

Our contributions are summarized as follows:
\begin{itemize}
    \item We propose \textbf{Curved Ray Expectation Positional Encoding}, a Unified Camera Model-compatible positional encoding that considers not only ray direction but also possible scene positions along the ray, enabling depth-aware positional encoding for curved-ray geometry.

    \item We apply CRePE to camera-conditioned video generation through a \textbf{Geometric Attention Adapter} for frozen video DiTs, incorporating token-wise scene distance into attention and stabilizing geometry prediction with pseudo radial-distance supervision.

    \item We provide controlled comparisons showing that CRePE exhibits a generally favorable performance trend over existing camera-conditioned baselines and a RayRoPE-style endpoint approximation, supported by supervision-free and zero-shot analyses, and further demonstrate that the same pathway can be extended through \textbf{Radial MixForcing} toward external radial-map control for scene-geometry-conditioned generation and source-video motion transfer.
\end{itemize}

\section{Related Work}
\label{sec:related_work}

\paragraph{Camera-conditioned video generation.}
Camera-conditioned video models control generated motion using pose embeddings, camera trajectories, rendered geometric guidance, or dedicated camera branches~\cite{he2025cameractrlenablingcameracontrol,wang2024motionctrlunifiedflexiblemotion,Yang_2024,zhang2024camerasraysposeestimation,bai2025recammaster}. ReCamMaster~\cite{bai2025recammaster} is a strong external baseline for trajectory-conditioned generation, but its formulation is tied to a pinhole setting. UCPE~\cite{zhang2025unified} instead conditions video diffusion with unified-camera rays and is therefore the closest baseline for our setting. CRePE builds on the UCPE interface but adds a token-wise ray-position variable inside attention. By contrast, depth-conditioned video generators inject dense structural features into a generative backbone~\cite{zhang2024controlvideo,lin2025ctrladapter,xing2024makeyourvideo,jiang2025vace,xi2026omnivdiff}.

\paragraph{Geometric positional encoding.}
Rotary positional encoding~\cite{su2023roformerenhancedtransformerrotary} makes attention depend on relative position. Multi-view variants replace image-grid position with camera-aware quantities: CaPE~\cite{kong2024eschernetgenerativemodelscalable}, GTA~\cite{miyato2024gtageometryawareattentionmechanism}, and PRoPE~\cite{li2025camerasrelativepositionalencoding} encode relative camera geometry; raymap methods concatenate ray origins or directions to features~\cite{sitzmann2022lightfieldnetworksneural,attal2022learningneurallightfields,sajjadi2022scenerepresentationtransformergeometryfree,gao2024cat3dcreate3dmultiview}; RayRoPE~\cite{wu2026rayropeprojectiveraypositional} predicts a token-wise ray-position interval under pinhole projection and applies expected RoPE after projection. CRePE shares this principle of averaging RoPE over ray-position uncertainty but differs from RayRoPE in two ways: it uses UCM projection for unified central cameras, and it integrates the phasor over a curved projected trajectory rather than a single endpoint-defined interval.

\paragraph{Central-camera geometry and radial distance.}
Wide-angle and fisheye videos are better described by central-camera models such as UCM than by pinhole intrinsics. Radial distance---distance from the camera center along the viewing ray---is a natural scalar for such cameras because it remains defined regardless of lens distortion. Recent geometric foundation models provide dense monocular scene geometry across diverse cameras~\cite{wang2024dust3rgeometric3dvision,leroy2024groundingimagematching3d,wang2025vggtvisualgeometrygrounded,piccinelli2025unik3duniversalcameramonocular}. We use these predictions only as pseudo radial-distance anchors for CRePE's geometry head.

\begin{table}[t]
\centering
\small
\setlength{\tabcolsep}{4pt}
\caption{\textbf{Geometry-aware positional encodings for camera-conditioned video DiTs.}
The RayRoPE column describes the original pinhole design; our $K=2$ endpoint ablation instead shares CRePE's UCM projection. ``--'' marks axes not applicable to a method without a learnable depth axis.}
\label{tab:method_comparison}
\begin{tabular*}{\linewidth}{@{\extracolsep{\fill}}lccc@{}}
\toprule
Property & UCPE~\cite{zhang2025unified} & RayRoPE~\cite{wu2026rayropeprojectiveraypositional} & CRePE (ours) \\
\midrule
Non-pinhole / UCM projection            & \cmark & \xmark & \cmark \\
Ray-position-adaptive phase             & \xmark & \cmark & \cmark \\
Curved projected-path integration       & --     & \xmark & \cmark \\
Pseudo-GT geometric grounding           & --     & \xmark & \cmark \\
Joint camera+depth control interface    & \xmark & \xmark & \cmark \\
\bottomrule
\end{tabular*}
\end{table}

\section{Method}
\label{sec:method}

\subsection{Overview}
\label{sec:method:overview}

\begin{figure}[t]
\centering
\includegraphics[width=0.9\linewidth]{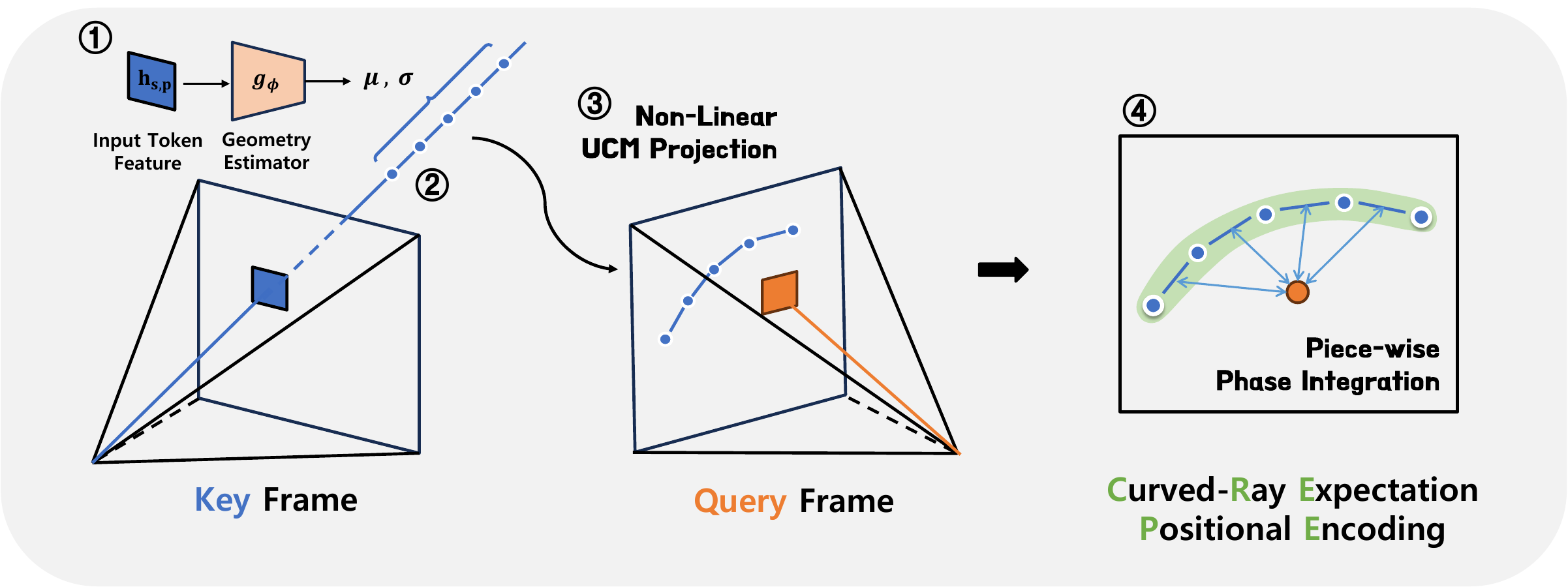}
\caption{\textbf{CRePE.}
A shared geometry head $g_\phi$ predicts a log-distance center $\mu$ and uncertainty $|\sigma|$ from each key-side token feature $\mathbf{h}_{s,p}$, defining $K$ breakpoints $\{r_k\}$ along the source viewing ray (\textcircled{\small 1}).
The breakpoints are unprojected to 3D via the source UCM model $\Phi^{\mathrm{UCM}}_{\xi_s}$ (\textcircled{\small 2}), transported into the query frame and forward-projected via $\Pi^{\mathrm{UCM}}_{\xi_q}$, tracing a curved trajectory $\{\tilde{\mathbf{x}}^{(k)}\}$ in the query view (\textcircled{\small 3}).
CRePE aggregates the expected RoPE phasor along this trajectory by analytically integrating each of the $K{-}1$ piecewise-linear segments (\textcircled{\small 4}), yielding a token-level modulation $\bar\rho_{f,c}$ that captures ray-position uncertainty under non-pinhole projection.}
\label{fig:method_main}
\end{figure}

We start from a frozen video diffusion transformer with a UCPE-style camera adapter. For each token at layer $\ell$, source frame $s$, and patch $p$, the adapter has feature $\bm{h}_{\ell,s,p}$. CRePE adds a geometry head $g_\phi$ that predicts a log radial-distance center $\mu$ and interval width $\sigma$:
\begin{equation}
    (\mu_{\ell,s,p}, \sigma_{\ell,s,p}) = g_\phi(\bm{h}_{\ell,s,p}).
    \label{eq:geom_head_main}
\end{equation}
These predictions define a token-wise radial-distance distribution. CRePE converts this distribution into RoPE coefficients by lifting radial-distance breakpoints along the source ray and projecting them into each query camera. The resulting coefficients modulate attention in a selected set of middle DiT layers. All other adapter layers keep UCPE's relative-ray modulation (Figure~\ref{fig:method_main}).

\subsection{Curved-Ray Expected RoPE}
\label{sec:method:crepe}

CRePE interprets the predicted interval as a uniform distribution in log radial distance. Specifically, for each source token we define $z = \log r \sim \mathcal{U}(\mu-|\sigma|,\mu+|\sigma|)$, where $r$ denotes the radial distance measured from the source camera center along the token's UCM ray. We then discretize this log-radial interval with $K$ uniformly spaced breakpoints and exponentiate them:
\begin{equation}
    z_k = \mu - |\sigma| + \frac{k-1}{K-1}(2|\sigma|), \qquad r_k = \exp(z_k), \qquad k=1,\ldots,K .
    \label{eq:breakpoints_main}
\end{equation}
Each $r_k$ is lifted along the source UCM ray, transformed into the query camera, and projected through the query UCM to obtain a query-side coordinate $\tilde{\mathbf{x}}^{(k)}$. For a fixed RoPE frequency $\omega_f$ and coordinate channel $c$, define $\theta^{(k)}_{f,c}=\omega_f\tilde{x}^{(k)}_c$. CRePE treats consecutive breakpoints as piecewise-linear phase segments and integrates the phasor analytically:
\begin{equation}
    \bar{c}_{f,c}=\frac{1}{K-1}\sum_{k=1}^{K-1}\frac{\sin\theta^{(k+1)}_{f,c}-\sin\theta^{(k)}_{f,c}}{\theta^{(k+1)}_{f,c}-\theta^{(k)}_{f,c}},\quad
    \bar{s}_{f,c}=\frac{1}{K-1}\sum_{k=1}^{K-1}\frac{\cos\theta^{(k)}_{f,c}-\cos\theta^{(k+1)}_{f,c}}{\theta^{(k+1)}_{f,c}-\theta^{(k)}_{f,c}} .
    \label{eq:phasor_main}
\end{equation}
The pair $(\bar{c}_{f,c},\bar{s}_{f,c})$ forms the rotary coefficient $\bar\rho_{f,c}=\bar{c}_{f,c}+i\,\bar{s}_{f,c}$ for that coordinate and frequency. Eq.~\eqref{eq:phasor_main} therefore approximates
$\mathbb{E}_{z \sim \mathcal{U}(\mu-|\sigma|,\mu+|\sigma|)}\big[\exp\!\big(i\omega_f \tilde{x}_c(z)\big)\big]$,
where $\tilde{x}(z)$ is the query-view projection of the 3D point obtained by lifting radial distance $r=\exp(z)$ along the source UCM ray. Importantly, the expectation is taken over log radial distance, not over image coordinates or metric distance directly. CRePE thus computes the expected phasor along the projected trajectory, rather than applying RoPE at a single averaged position $\exp(i\omega_f \mathbb{E}[\tilde{x}_c])$. When $K=2$, the path degenerates to an endpoint interval; with $K>2$, CRePE captures non-affinity of the projected phase, including the curvature induced by UCM projection (Appendix~\ref{app:curved_path_protocols}). We use $K=5$ by default.
Implementation details, including the UCM projection equations, bounded query-view coordinates, patch-offset rays, and numerical handling of small phase differences, are given in Appendices~\ref{app:rope_ucm} and~\ref{app:crepe_details}.

\paragraph{CRePE inside attention.}
We apply the expected phasor in the geometric attention branch, following the relative-transform pattern of the UCPE adapter (Figure~\ref{fig:supp_architecture}). Let $i=(q,p_q)$ denote a query token in query frame $q$, and let $j=(s,p_s)$ denote a key/value token in source frame $s$. For a CRePE-enabled layer $\ell$, the geometry head predicts $(\mu_{\ell,s,p_s},\sigma_{\ell,s,p_s})$ from the key-side feature $\bm{h}_{\ell,s,p_s}$. Given the relative camera transform from source frame $s$ to query frame $q$, Eqs.~\eqref{eq:breakpoints_main}--\eqref{eq:phasor_main} produce a query-conditioned expected rotary coefficient $\bar\rho_{\ell,q\leftarrow s,p_s}=\{(\bar{c}_{f,c},\bar{s}_{f,c})\}_{f,c}$. We convert this coefficient into the corresponding block-diagonal rotary operator $\bar{R}_{j}\equiv\bar{R}_{\ell,q\leftarrow s,p_s}$; the query token uses its own coefficient in its own frame, $\bar{R}_{i}\equiv\bar{R}_{\ell,q\leftarrow q,p_q}$. Queries, keys, values, and outputs are then transformed as
\begin{equation}
\begin{gathered}
Q_{\ell,i} = \bar{R}_{i}^{\top} W_Q \bm{h}_{\ell,i}, \qquad
K_{\ell,j} = \bar{R}_{j}^{\top} W_K \bm{h}_{\ell,j}, \qquad
V_{\ell,j} = \bar{R}_{j}^{\top} W_V \bm{h}_{\ell,j},\\
A^{\mathrm{crepe}}_{\ell,i,j}=\operatorname{softmax}_{j}\!\left(\frac{Q_{\ell,i}^{\top}K_{\ell,j}}{\sqrt{d}}\right),
\qquad
O^{\mathrm{crepe}}_{\ell,i}=\bar{R}_{i}\sum_j A^{\mathrm{crepe}}_{\ell,i,j} V_{\ell,j}.
\end{gathered}
\label{eq:crepe_attention}
\end{equation}
Each logit thus depends on $\bar\rho_{\ell,q\leftarrow q,p_q}\bar\rho^{*}_{\ell,q\leftarrow s,p_s}$, the relative expected phase between the query and the key in the query view. For a unit-magnitude phasor, $\bar{R}^{\top}=\bar{R}^{-1}$, which recovers the $D^{\top}$/$D^{-1}$/$D$ pattern of Figure~\ref{fig:supp_architecture}; because $|\bar\rho_{f,c}|\le1$, we use the transpose.
The output of this branch is passed through the same zero-initialized output projection used by the adapter and added residually to the frozen DiT feature.
CRePE is used only in the selected middle layers; the remaining layers keep the UCPE relative-ray modulation. This preserves the stable ray-only camera signal while injecting depth-aware projected-path modulation where radial-distance information is most recoverable.

\subsection{Geometric Attention Adapter and Radial-Distance Supervision}
\label{sec:method:gaa}

The geometry head in Eq.~\eqref{eq:geom_head_main} affects attention directly, so incorrect predictions can become a harmful positional shortcut. We therefore supervise it with pseudo radial-distance maps from a monocular geometry foundation model, UniK3D~\cite{piccinelli2025unik3duniversalcameramonocular}. We first apply validity filtering to the raw metric pseudo distance $\hat{r}^{\mathrm{m}}$: a target is invalid if UniK3D fails, is marked unreliable by our validity filters, returns a non-finite or non-positive value, falls outside the supported distance range, or predicts $\hat{r}^{\mathrm{m}}>r_{\max}=20\,\mathrm{m}$. Invalid or far-field values are not clipped into pseudo ground truth. Only valid metric targets are normalized by a per-clip near-distance statistic, pooled to the token grid, and used as normalized targets $\hat{r}$ in the loss. From $(\mu,\sigma)$ we decode the predicted normalized radial distance $\bar{r}=\exp(\mu)$ and an uncertainty scale $s$ derived from the log interval. The auxiliary loss and the full objective are
\begin{equation}
    \mathcal{L}_{\mathrm{rad}} =
    \frac{1}{|\mathcal{T}|}\sum_{n\in\mathcal{T}}
    \left(\frac{|\bar{r}_{n}-\hat{r}_{n}|}{s_{n}} + \alpha\log s_{n}\right),
    \qquad
    \mathcal{L}=\mathcal{L}_{\mathrm{diff}}+\lambda_{\mathrm{rad}}\mathcal{L}_{\mathrm{rad}},
    \label{eq:rad_loss_main}
\end{equation}
where $\mathcal{T}$ is the valid token set indexed by $n$, $\alpha=1$, $\mathcal{L}_{\mathrm{diff}}$ is the backbone diffusion objective, and $\lambda_{\mathrm{rad}}=10^{-3}$. We supervise the geometry head at a single selected layer, which trains the head shared across all CRePE-modulated layers, and keep the video DiT backbone frozen. Only $0.061\%$ of pixels are true UniK3D failures; after the $20\,$m far-field cut, $56.3\%$ of pixels ($55.7\%$ of tokens) remain valid, and the excluded rays (mostly sky) use the model's own prediction. In practice, this radial-distance anchor is important: diffusion-only CRePE can learn plausible ray-position maps early, but degrades under longer optimization (Section~\ref{sec:experiments:robustness}).

\subsection{Radial MixForcing}
\label{sec:method:radial_mixforcing}

Radial MixForcing is a training-time teacher-substitution algorithm for the CRePE ray-position pathway. Its purpose is to teach the adapter to accept externally supplied radial maps while preserving a safe camera-only fallback. Let $(\mu_{\mathrm{pred}},\sigma_{\mathrm{pred}})$ be the geometry-head output, let $\hat{r}^{\mathrm{m}}$ be the raw metric pseudo radial distance, let $\hat{r}$ be its per-clip-normalized value when valid, let $\mu_{\mathrm{gt}}=\log \hat{r}$, and let $M$ be a sampled teacher-substitution mask. We define pseudo-GT validity on the raw metric output as $V_{\mathrm{gt}}$ below; distances above the validity threshold are therefore treated as unknown, not as valid targets clipped to $20\,\mathrm{m}$. CRePE consumes the effective radial distribution
\begin{equation}
    V_{\mathrm{gt}}=\mathbf{1}\!\left[\mathrm{valid}_{\mathrm{UniK3D}}(\hat{r}^{\mathrm{m}})\wedge 0<\hat{r}^{\mathrm{m}}\le r_{\max}\right],
    \qquad
    (\mu_{\mathrm{eff}},\sigma_{\mathrm{eff}})=
    \begin{cases}
    (\mu_{\mathrm{gt}},\sigma_T) & \text{if } M\wedge V_{\mathrm{gt}},\\
    (\mu_{\mathrm{pred}},\sigma_{\mathrm{pred}}) & \text{otherwise},
    \end{cases}
    \label{eq:mix_effective_main}
\end{equation}
where $\sigma_T$ is a narrow teacher interval. The second branch covers every invalid ray: no teacher geometry is injected and that ray uses the model prediction. The auxiliary loss in Eq.~\eqref{eq:rad_loss_main} is still applied only on valid tokens, so the head learns from reliable pseudo radial distances while invalid regions remain self-predicted. At inference, an external radial map follows the same rule after validity filtering and scale normalization: valid external rays may replace the predicted tensor, while invalid or missing rays fall back to $(\mu_{\mathrm{pred}},\sigma_{\mathrm{pred}})$. We use the supervised CRePE model for the main camera-only benchmark and the MixForcing-trained model (CRePE-Mix) for external-radial-map control. The schedule and teacher-width details are in Appendix~\ref{app:radial_mixforcing}; an inference-time sweep of $\sigma_T$ (trained at $0.1$) over a $20\times$ range changes CamMC by only $3.9\%$ (Appendix~\ref{app:zeroshot_external}).

\subsection{Layer Placement}
\label{sec:method:placement}

Applying CRePE to every layer is unnecessary and can degrade visual quality. We probe the frozen DiT layers for recoverable radial-distance information and, with a placement ablation under the reduced-budget protocol of Section~\ref{sec:ablation_num_points_with_pose} (1K steps, $\lambda_{\mathrm{rad}}=0$, pose stride 4), find that middle layers provide the best tradeoff between geometric control and generation quality. Our default model applies CRePE to the middle 10 blocks and retains UCPE modulation elsewhere. The probing curve and placement ablation are in Appendix~\ref{app:layer_ablation}.

\section{Experiments}
\label{sec:experiments}

\subsection{Setup}
\label{sec:experiments:setup}

\paragraph{Backbone and training.}
We use Wan2.1-T2V-1.3B~\cite{wan2025wan} as the video diffusion backbone and keep its weights frozen. CRePE is trained through the camera-conditioning adapter for 10K steps using AdamW. The default setting applies CRePE to the middle 10 DiT blocks, uses $K=5$ radial-distance breakpoints, and supervises the shared geometry head at a single selected layer. Full hyperparameters are in Appendix~\ref{app:crepe_details}.

\paragraph{Dataset.}
We train and evaluate on PanShot~\cite{zhang2025unified}, a multi-camera video dataset covering panning shots with pinhole, wide-angle, and fisheye camera settings, using its released train/test split. Each clip has a camera trajectory and UCM lens parameters. We evaluate all 340 test clips without filtering by camera-motion magnitude; reduced-budget ablations use a fixed 50-clip subset of the test split (13 pinhole, 37 non-pinhole). Pseudo radial-distance maps are produced offline with UniK3D~\cite{piccinelli2025unik3duniversalcameramonocular} and used only for training supervision or optional external-radial-map interventions. Dataset details are in Appendix~\ref{app:dataset_details}.

\paragraph{Metrics.}
We evaluate camera controllability with rotation error (RotErr), scale-aligned translation error (TransErr), and CamMC, recovering relative pose from generated videos with ViPE~\cite{huang2025vipe} following UCPE~\cite{zhang2025unified}. We evaluate generation quality with FVD, FVD-C, FID, Inception Score, VBench~\cite{huang2024vbench}, and CLIP-based image/text scores. Lens and orientation diagnostics, estimated with GeoCalib~\cite{veicht2024geocalib}, include radial-distortion-coefficient errors $(k_1,k_2)$ and pitch, gravity, and up-vector errors. Main comparisons recover pose at frame stride 1; the reduced-budget ablations and zero-shot studies use stride 4, so absolute pose values are comparable only within a protocol.

\subsection{Baselines}
\label{sec:experiments:baselines}

The main evaluation compares against \textbf{ReCamMaster}~\cite{bai2025recammaster} and \textbf{UCPE}~\cite{zhang2025unified}. ReCamMaster is the external camera-controlled generation baseline; UCPE is the controlled baseline with the same Wan backbone and adapter interface but without CRePE or radial-distance supervision. We evaluate a \textbf{RayRoPE-style endpoint PE}~\cite{wu2026rayropeprojectiveraypositional} in a separate positional-encoding ablation, where it isolates the difference between endpoint-based expected RoPE and CRePE's UCM-compatible curved-ray integration; it uses CRePE's UCM projection with $K=2$ and is not the full RayRoPE system.

\subsection{Main Camera-Controlled Generation Results}
\label{sec:experiments:main_results}

\begin{figure}[t]
\centering
\includegraphics[width=0.75\linewidth]{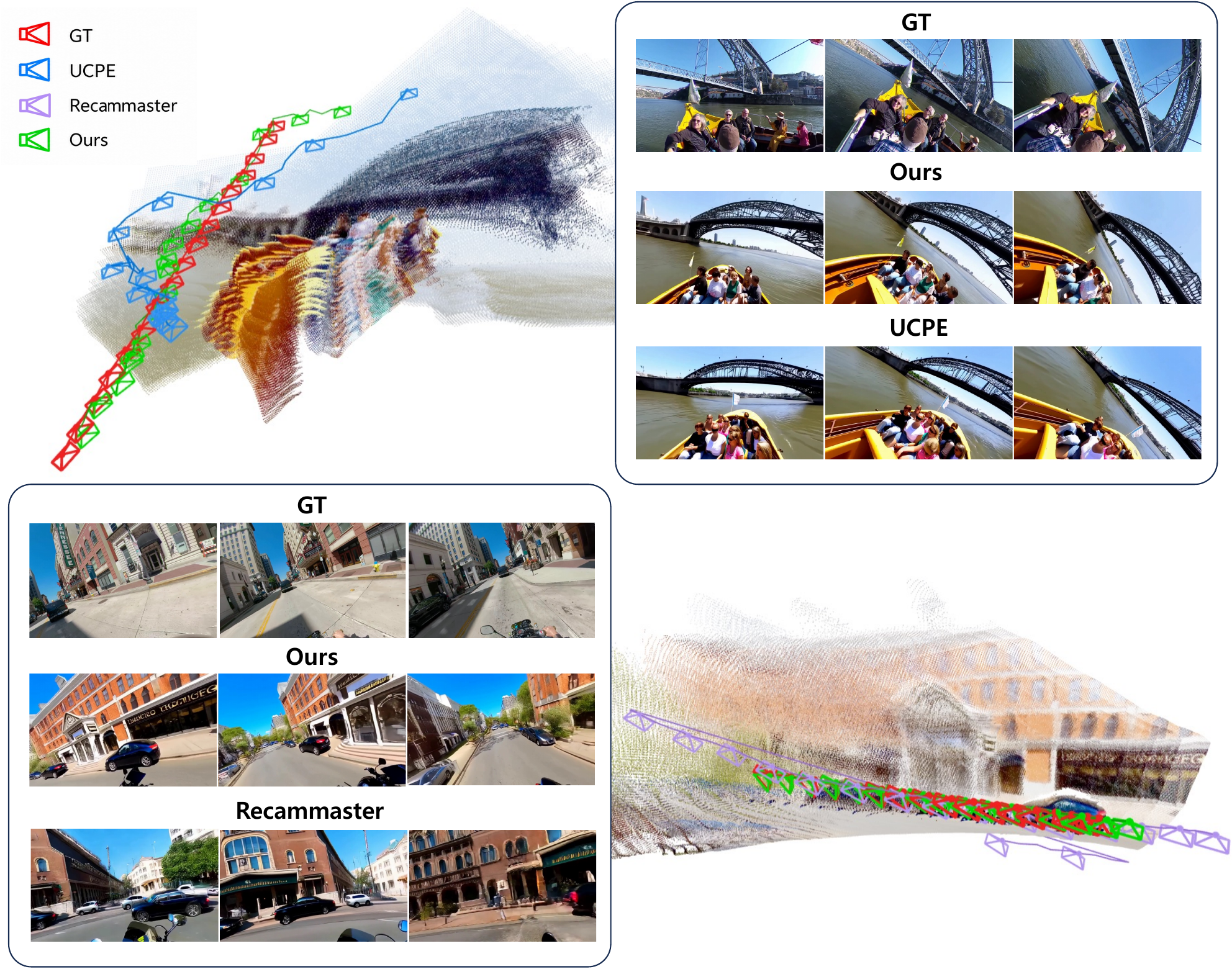}
\caption{\textbf{Qualitative camera-control comparison.} Generated videos under matched prompts and camera trajectories. The main comparison includes ReCamMaster, UCPE, and CRePE. CRePE follows the requested camera trajectory most stably under non-pinhole lenses.}
\label{fig:qual_main}
\end{figure}

Table~\ref{tab:pinhole_nonpinhole_main} and Figure~\ref{fig:qual_main} report the main comparison across lens families. CRePE improves the overall camera-conditioning profile relative to the baselines on both pinhole and non-pinhole subsets. Compared with UCPE, CRePE improves most metrics, while UCPE remains a strong baseline and is slightly better on FVD (463.5 vs.\ 465.3), FVD-C, and IS (Table~\ref{tab:overall_full_comparison}); we therefore describe CRePE's video quality as competitive rather than improved. This tradeoff suggests that CRePE improves scene-geometry-aware camera following, but the direct ray/lens signal in UCPE is still useful.

\begin{table}[t]
\centering
\setlength{\tabcolsep}{3.5pt}
\caption{\textbf{Pinhole and non-pinhole camera-generation comparison} (340 PanShot test clips; pose stride 1).
We report representative metrics across pose control, text/image alignment, lens distortion, orientation, and VBench quality.
Non-pinhole and pinhole subsets are grouped by rows for readability.
CRePE consistently improves over UCPE and ReCamMaster on the representative camera-control, geometry-aware, and VBench quality metrics.}
\label{tab:pinhole_nonpinhole_main}
\resizebox{\textwidth}{!}{%
\begin{tabular}{llrrrrrrrr}
\toprule
\multirow{2}{*}{Subset}
& \multirow{2}{*}{Method}
& \multicolumn{1}{c}{Pose}
& \multicolumn{2}{c}{Text/Image}
& \multicolumn{2}{c}{Lens}
& \multicolumn{1}{c}{Orientation}
& \multicolumn{2}{c}{VBench} \\
\cmidrule(lr){3-3}
\cmidrule(lr){4-5}
\cmidrule(lr){6-7}
\cmidrule(lr){8-8}
\cmidrule(lr){9-10}
&
& CamMC $\downarrow$
& CS-I $\uparrow$
& CS-T $\uparrow$
& $k_1$ $\downarrow$
& $k_2$ $\downarrow$
& Gravity $\downarrow$
& Aes. Qual. $\uparrow$
& BG Cons. $\uparrow$ \\
\midrule
\multirow{3}{*}{Non-pinhole}
& ReCamMaster & 54.56 & 98.69 & 24.60 & 0.186 & 0.154 & 11.19 & 0.5596 & 0.9484 \\
& UCPE        & 20.60 & 98.80 & 24.78 & 0.183 & 0.154 &  7.08 & 0.5616 & 0.9537 \\
& \textbf{CRePE} & \textbf{19.02} & \textbf{98.81} & \textbf{24.80} & \textbf{0.157} & \textbf{0.116} & \textbf{6.27} & \textbf{0.5639} & \textbf{0.9542} \\
\midrule
\multirow{3}{*}{Pinhole}
& ReCamMaster & 57.83 & 98.65 & 25.58 & 0.179 & 0.152 & 10.98 & 0.5605 & 0.9441 \\
& UCPE        & 18.50 & 98.59 & 25.86 & 0.175 & 0.151 &  7.10 & 0.5619 & 0.9457 \\
& \textbf{CRePE} & \textbf{15.33} & \textbf{98.65} & \textbf{25.93} & \textbf{0.153} & \textbf{0.113} & \textbf{6.13} & \textbf{0.5620} & \textbf{0.9471} \\
\bottomrule
\end{tabular}%
}
\end{table}

\subsection{Positional-Encoding Ablation: RayRoPE-Style Endpoint versus CRePE}
\label{sec:ablation_num_points_with_pose}

We use the RayRoPE-style endpoint PE as a controlled ablation of the positional encoding rather than as a main camera-generation baseline.
It corresponds to a two-point endpoint approximation of the ray-position interval ($K=2$), while CRePE performs curved-ray phasor integration by sampling multiple radial-distance breakpoints along the projected trajectory.
This comparison isolates whether explicitly integrating over the curved ray path provides benefits beyond the endpoint-based approximation.
All variants use the same 1K-step diffusion-only objective ($\lambda_{\mathrm{rad}}=0$), parameter count, and data, with pose evaluated at stride 4; UniK3D is used neither in training nor at inference, and interior breakpoints are deterministic evaluations of the same predicted $(\mu,\sigma)$.

Table~\ref{tab:ablation_num_points_with_pose} shows that CRePE consistently improves over the endpoint approximation in overall average rank.
$K=1$ (a single point) collapses in quality (FID 242.7), and the largest pose gain comes from $K=2$ to $K=3$ (CamMC $14.75\rightarrow12.31$).
With $K=5$, CRePE achieves the best balance across pose accuracy, text/image alignment, video quality, lens distortion, and orientation metrics.
Increasing the number of samples to $K=7$ slightly improves some distortion-related metrics, but does not further improve the overall balance.
We therefore use $K=5$ as our default setting.
These gains are not a sample-count effect: for an affine phase path the coefficient is independent of $K$, and the gains appear on both pinhole and non-pinhole clips, so they are not purely due to UCM curvature (Appendix~\ref{app:curved_path_protocols}).

\begin{table}[t]
\centering
\small
\setlength{\tabcolsep}{5.5pt}
\caption{\textbf{Sampling-point ablation for CRePE} (1K steps, $\lambda_{\mathrm{rad}}=0$, 50-clip subset, pose stride 4).
We compare the RayRoPE-style endpoint PE ($K=2$) and CRePE variants using representative metrics.
Avg.\ Rank is computed among the five settings over 19 metrics, excluding IS Std.
CRePE with $K=5$ achieves the best overall balance, while $K=3$ leads pose.}
\label{tab:ablation_num_points_with_pose}
\begin{tabular}{lrrrrrr}
\toprule
Setting & Avg. Rank $\downarrow$ & CamMC $\downarrow$ & Rot. $\downarrow$ & Gravity $\downarrow$ & CS-T $\uparrow$ & FID $\downarrow$ \\
\midrule
$K=1$ (point) & 4.53 & 17.64 & 6.73 & 18.22 & 20.57 & 242.7 \\
$K=2$ (endpoint PE) & 3.27 & 14.75 & 5.65 & 10.46 & 25.75 & \textbf{135.3} \\
$K=3$ (CRePE) & 2.94 & \textbf{12.31} & \textbf{4.43} & 9.90 & 25.76 & 139.1 \\
$K=5$ (CRePE, default) & \textbf{1.71} & 13.45 & 4.65 & \textbf{9.69} & \textbf{25.76} & 136.6 \\
$K=7$ (CRePE) & 2.56 & 14.23 & 4.70 & 9.80 & 25.70 & 135.5 \\
\bottomrule
\end{tabular}
\end{table}

\subsection{Supervision, Robustness, and Zero-Shot Transfer}
\label{sec:experiments:robustness}

\paragraph{Geometry as phase, not extra information.}
A matched Depth-Concat control predicts the same pseudo radial signal under the same loss and 10K-step schedule, but concatenates it to every token through a projection MLP ($+6$M parameters). It reaches comparable quality (FVD 473.0 vs.\ 465.3) yet trails even depth-free UCPE on pose (Table~\ref{tab:rebuttal_evidence}A), so among the tested alternatives the advantage comes from making radial distance a query-dependent phase, not from access to monocular geometry.

\begin{table}[t]
\centering
\caption{\textbf{Controls and zero-shot transfer.} Pose stride 1 in A/B (B: 1K runs on the 50-clip subset, 10K runs on all 340 clips), 4 in C. Lower is better except VBench-OC (overall consistency).}
\label{tab:rebuttal_evidence}
\label{tab:app_zeroshot_full}
\footnotesize
\begin{minipage}[t]{0.48\linewidth}
\centering
\textbf{A. Geometry injection (10K steps)}\par
\setlength{\tabcolsep}{1.5pt}
\begin{tabular*}{\linewidth}{@{\extracolsep{\fill}}lrrrr@{}}
\toprule
Method & CamMC & Rot. & Trans. & FVD \\
\midrule
UCPE (no depth) & 21.87 & 6.66 & 17.54 & \textbf{463.5} \\
Depth-Concat & 23.21 & 7.48 & 18.11 & 473.0 \\
CRePE (phase) & \textbf{18.15} & \textbf{6.33} & \textbf{13.78} & 465.3 \\
\bottomrule
\end{tabular*}
\end{minipage}\hfill
\begin{minipage}[t]{0.49\linewidth}
\centering
\textbf{B. Supervision and training duration}\par
\setlength{\tabcolsep}{1.5pt}
\begin{tabular*}{\linewidth}{@{\extracolsep{\fill}}lrrrr@{}}
\toprule
Metric & 1K no & 1K sup & 10K no & 10K sup \\
\midrule
CamMC & \textbf{52.02} & 56.27 & 22.19 & \textbf{18.15} \\
RotErr & \textbf{18.00} & 18.21 & 8.11 & \textbf{6.33} \\
SSI-L1 & 0.129 & \textbf{0.117} & 0.133 & \textbf{0.087} \\
\bottomrule
\end{tabular*}
\end{minipage}

\smallskip
\textbf{C. Zero-shot transfer without tuning}\par
\setlength{\tabcolsep}{2pt}
\begin{tabular*}{\linewidth}{@{\extracolsep{\fill}}lrrrrrrrr@{}}
\toprule
\multirow{2}{*}{Method} & \multicolumn{4}{c}{KITTI-360 (32 scenes, $180^\circ$ fisheye)} & \multicolumn{4}{c}{RealCam-Vid (100 scenes)} \\
\cmidrule(lr){2-5}\cmidrule(lr){6-9}
& CamMC & Rot. & Trans. & VBench-OC & CamMC & Rot. & Trans. & VBench-OC \\
\midrule
UCPE & 22.32 & 16.20 & 9.34 & 0.2097 & 12.17 & 7.02 & 7.00 & 0.1894 \\
CRePE (no-sup) & 18.18 & 14.01 & \textbf{5.15} & \textbf{0.2198} & 12.31 & 6.68 & 7.22 & 0.2146 \\
CRePE & \textbf{17.01} & \textbf{12.70} & 5.59 & 0.2137 & \textbf{9.55} & \textbf{5.79} & \textbf{4.89} & \textbf{0.2148} \\
\bottomrule
\end{tabular*}
\end{table}

\paragraph{Supervision anchors long training.}
At 1K steps, $\mathcal{L}_{\mathrm{rad}}$ provides no camera-control advantage, so the supervision-free protocol of Section~\ref{sec:ablation_num_points_with_pose} isolates the formulation. At 10K steps (full test split; the 1K runs use the 50-clip subset), the unsupervised head's radial self-consistency error (SSI-L1 against depth re-estimated from its own generated video) is 0.133 versus 0.087 with supervision (0.129 vs.\ 0.117 at 1K), and the CamMC ordering reverses, with the unsupervised model falling below UCPE (Table~\ref{tab:rebuttal_evidence}B). Few monocular models provide metric radial distance under the UCM, which motivated UniK3D; UniDepthV2 supervision~\cite{piccinelli2026unidepthv2} on the pinhole subset (1K steps, pose stride 4) is comparable to UniK3D (CamMC 15.41 vs.\ 15.23), suggesting that the results are not specific to one teacher.

\paragraph{Noisy dynamic regions.}
On 50 PanShot clips with moving-object masks~\cite{huang2025segmentmotion} (34 with valid support; one middle-window layer at one denoising step; Appendix~\ref{app:supervision_robustness}), dynamic tokens disagree more with the pseudo target (log-MAE 0.294 vs.\ 0.208) but receive $1.51\times$ wider predicted uncertainty than static tokens. Across frames, CRePE's median log-radial statistic changes by 0.0156 versus 0.0375 for UniK3D and 0.0084 for the SLAM-consistent ViPE reference~\cite{huang2025vipe}, and CRePE tracks ViPE's scale changes more closely (drift disagreement 0.020 vs.\ 0.034; favoring CRePE on $82\%$ of supported clips). CRePE thus filters rather than copies teacher jitter; its latent is not metric moving-object depth.

\paragraph{Generalization beyond PanShot.}
Without tuning, CRePE improves every pose metric over UCPE on 32 real $180^\circ$ fisheye KITTI-360 scenes~\cite{liao2023kitti360} and on 100 RealCam-Vid~\cite{zheng2025realcamvid} scenes drawn from RealEstate10K~\cite{zhou2018stereo}, DL3DV~\cite{ling2024dl3dv}, and MiraData~\cite{ju2024miradata} (Table~\ref{tab:rebuttal_evidence}C), reducing CamMC by $24\%$ and $22\%$, respectively, and also improves VBench overall consistency. The no-supervision model already transfers on KITTI-360, while supervision provides the margin on diverse pinhole videos.

\begin{figure}[t]
\centering
\includegraphics[width=0.75\linewidth]{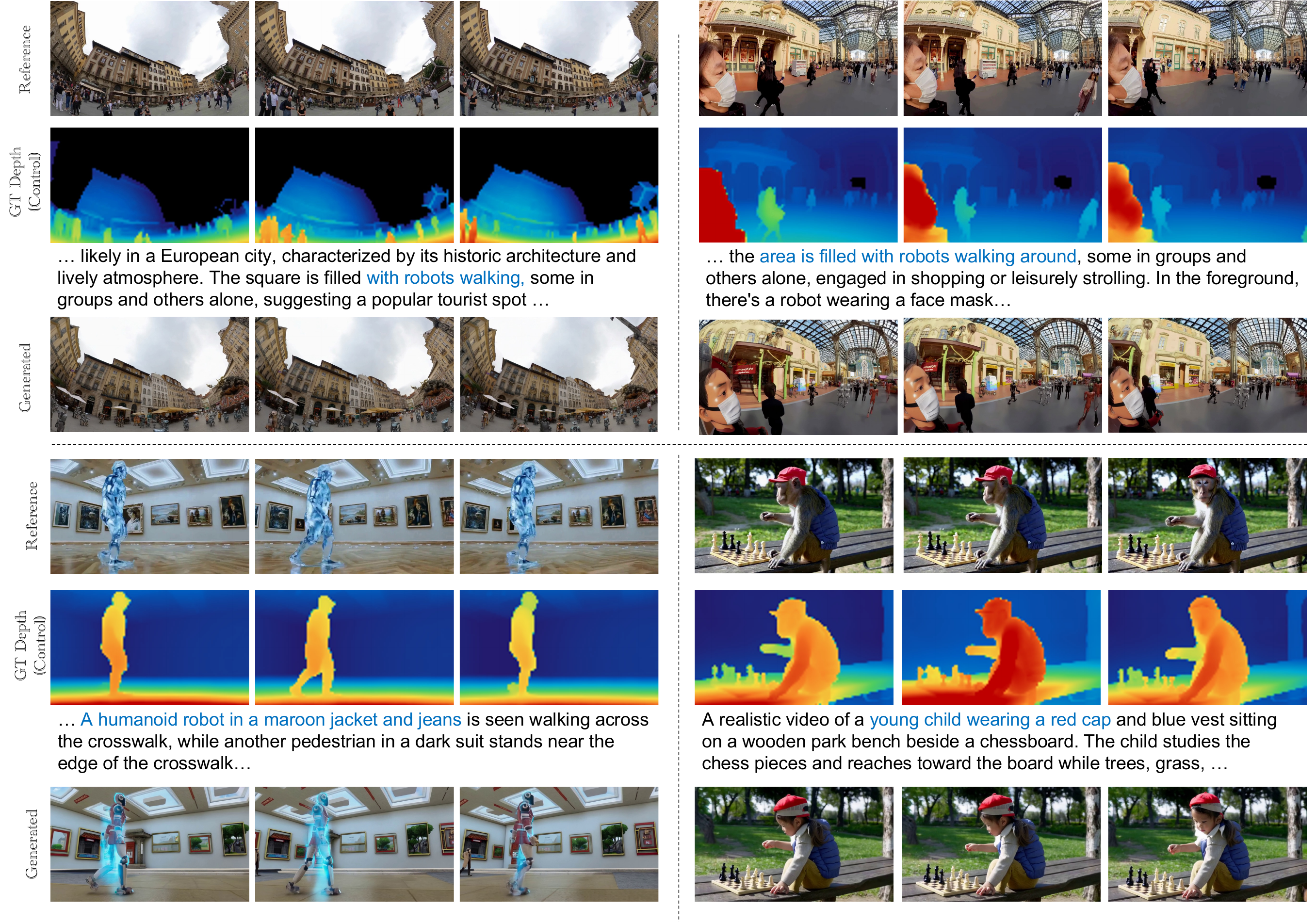}
\caption{\textbf{Qualitative external-radial-map control.} CRePE-Mix can consume externally supplied radial-distance maps through the same positional-encoding pathway used during camera-conditioned generation. We show scene-geometry-conditioned generation and source-video motion transfer qualitative results on PanShot (top) and in-the-wild AI-generated videos (bottom). ``GT Depth (Control)'' marks supplied pseudo radial maps, not sensor ground truth. The motion-transfer examples (bottom) start denoising from the source video's latents noised to $75\%$ (SDEdit-style~\cite{meng2022sdedit}) rather than from pure noise, so source appearance can also carry over; the quantitative adherence in Table~\ref{tab:mix_quant} is measured from pure noise.}
\label{fig:external_radial}
\end{figure}

\subsection{External Radial-Map Conditioning and Motion Transfer}
\label{sec:experiments:external_radial}

CRePE exposes the ray-position variable through the positional-encoding pathway. Although our primary benchmark is camera-conditioned generation, this interface also allows intervention: at inference, CRePE-Mix can replace the predicted radial-distance tensor with an external radial map after the same validity filtering and scale normalization used for pseudo targets. We show two qualitative uses in Figure~\ref{fig:external_radial}. First, the same prompt and camera trajectory can be generated with different radial maps, changing scene layout through the PE pathway. Second, a source video can provide a sequence of radial maps, allowing motion and scene-geometry transfer while using a new text prompt and target camera trajectory. Invalid or missing external radial values do not overwrite the predicted tensor; those rays continue to use the model prediction according to Eq.~\eqref{eq:mix_effective_main}.

\begin{figure}[t]
\centering
\begin{minipage}[t]{0.47\linewidth}
\vspace{0pt}
\centering
\includegraphics[width=\linewidth]{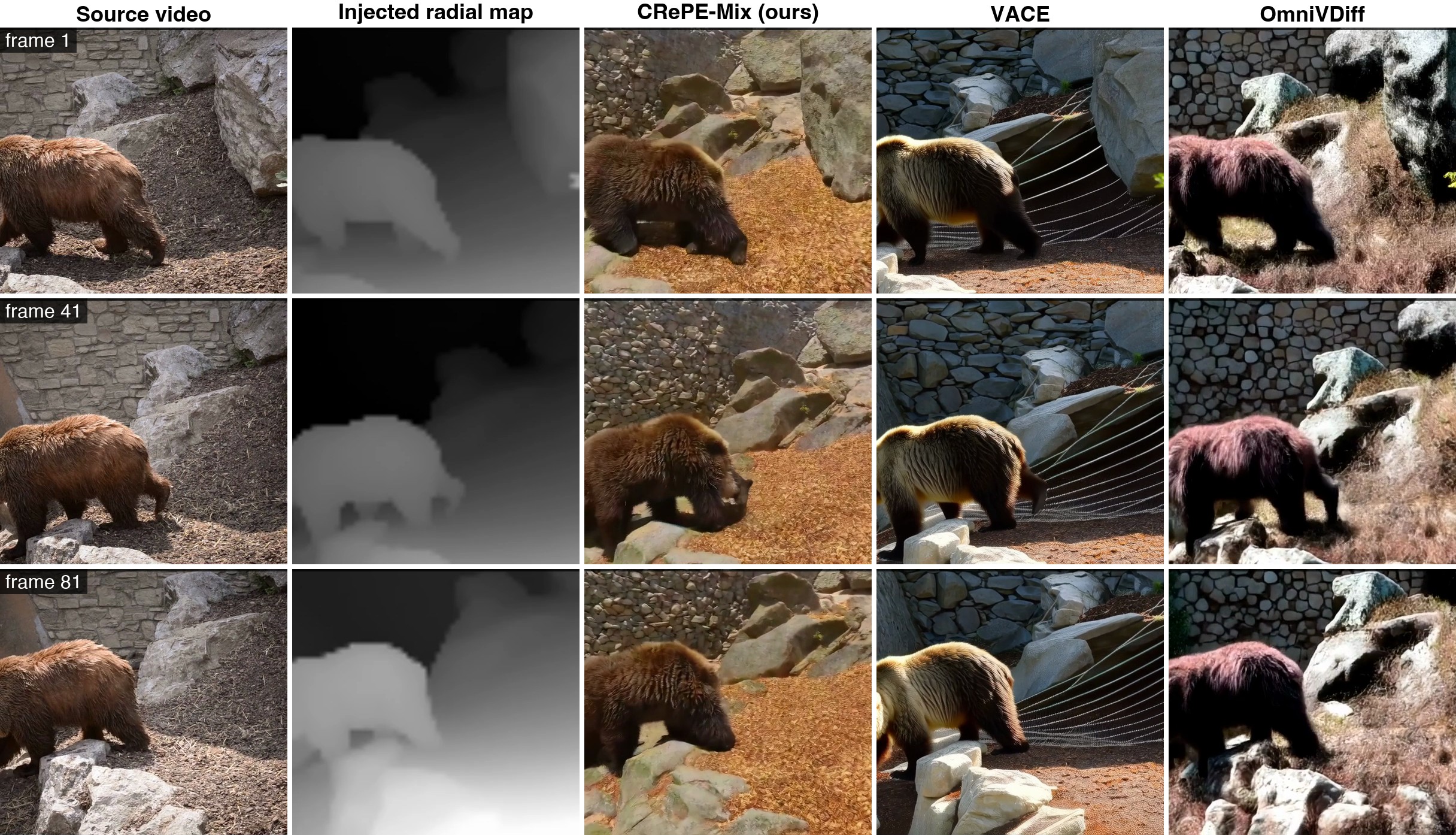}
\raggedright
\caption{\textbf{DAVIS radial-map control.}
Columns: source video, UniK3D control, CRePE-Mix, VACE, OmniVDiff; rows: frames 1/41/81. Black control regions are invalid far-field rays, where CRePE-Mix uses its own prediction.}
\label{fig:davis_qual}
\end{minipage}\hfill
\begin{minipage}[t]{0.51\linewidth}
\vspace{0pt}
\captionof{table}{\textbf{DAVIS radial-map adherence} (33 sequences). Cells: pixel (top) / token (bottom); best in bold. $\delta_1$ uses mean-scale alignment; Jitter: median frame-to-frame change of the fitted depth scale; UDv2: UniDepthV2 inputs.}
\label{tab:mix_quant}
\centering
\scriptsize
\setlength{\tabcolsep}{2pt}
\newcommand{\ptscore}[2]{\shortstack{#1\\#2}}
\begin{tabular}{@{}lcccc@{}}
\toprule
Method & SILog $\downarrow$ & AbsRel $\downarrow$ & $\delta_1\uparrow$ & Jitter $\downarrow$ \\
\midrule
ControlVideo~\cite{zhang2024controlvideo} & \ptscore{0.3952}{0.3842} & \ptscore{0.4018}{0.3873} & \ptscore{0.5465}{0.5486} & \ptscore{0.2739}{0.2751} \\[1pt]
Ctrl-Adapter~\cite{lin2025ctrladapter} & \ptscore{0.4942}{0.4854} & \ptscore{0.6305}{0.6150} & \ptscore{0.4437}{0.4476} & \ptscore{0.4013}{0.4023} \\[1pt]
Make-Your-Video~\cite{xing2024makeyourvideo} & \ptscore{0.5942}{0.5892} & \ptscore{0.8540}{0.8426} & \ptscore{0.3140}{0.3182} & \ptscore{0.5087}{0.5089} \\[1pt]
OmniVDiff~\cite{xi2026omnivdiff} & \ptscore{0.3764}{0.3701} & \ptscore{0.3902}{0.3842} & \ptscore{0.5096}{0.5133} & \ptscore{0.3184}{0.3181} \\[1pt]
VACE~\cite{jiang2025vace} & \ptscore{\textbf{0.2982}}{0.2877} & \ptscore{0.3016}{0.2929} & \ptscore{0.6220}{0.6259} & \ptscore{0.2518}{0.2521} \\[1pt]
CRePE-Mix (UDv2) & \ptscore{0.3218}{0.3059} & \ptscore{0.3307}{0.3161} & \ptscore{0.6348}{0.6386} & \ptscore{0.2174}{0.2164} \\[1pt]
\textbf{CRePE-Mix} & \ptscore{0.2990}{\textbf{0.2834}} & \ptscore{\textbf{0.2719}}{\textbf{0.2578}} & \ptscore{\textbf{0.6400}}{\textbf{0.6439}} & \ptscore{\textbf{0.1781}}{\textbf{0.1783}} \\
\bottomrule
\end{tabular}
\end{minipage}
\end{figure}

Beyond these qualitative examples, we quantify adherence on 33 DAVIS-2016 sequences~\cite{perazzi2016davis}: all methods receive identical UniK3D source maps, and geometry re-estimated from each generated video is compared at pixel and $16\!\times\!16$-pooled token resolution. CRePE-Mix is best on all token-level metrics and on seven of eight pixel/token comparisons (Table~\ref{tab:mix_quant}; within 0.0008 of VACE on pixel SILog), with about $30\%$ lower Scale Jitter than the strongest baseline, despite modifying only positional coefficients. On 48 valid non-pinhole PanShot scenes, it also leads VACE and OmniVDiff in AbsRel/$\delta_1$/Scale Jitter (0.550/0.381/0.088 vs.\ 0.833/0.371/0.095 and 0.929/0.097/0.233). In Figure~\ref{fig:davis_qual}, CRePE-Mix reproduces the source layout and motion, whereas VACE adds structure absent from the control map and OmniVDiff drifts in appearance.

\subsection{Limitations}
\label{sec:limitations}

CRePE improves geometry-aware camera conditioning, but does not solve all camera-control challenges.
Camera recovery from generated videos is itself noisy, and radial adherence is re-estimated with the same UniK3D model that supplies pseudo labels, so pose and geometry metrics should be interpreted together with qualitative examples; main-table values are single-run point estimates.
The radial-distance supervision depends on the quality and scale normalization of monocular geometry predictions, which may introduce errors when the estimated geometry is unreliable or poorly calibrated, as on moving objects.
In addition, CRePE currently focuses on central-camera models represented by UCM, and extending the formulation to more general non-central or rolling-shutter camera models remains future work.
Finally, external-radial-map baselines use different backbones, so we do not claim state-of-the-art depth conditioning; a motion-transfer benchmark is future work.

\section{Conclusion}
\label{sec:conclusion}

We presented CRePE, a curved-ray expectation positional encoding for unified-camera-controlled video generation. CRePE represents each token as a depth-aware positional distribution along its source ray and integrates the expected RoPE phasor along the UCM-projected curved trajectory, extending ray-only camera conditioning with token-wise scene-geometry awareness while remaining compatible with non-pinhole lenses. A Geometric Attention Adapter realizes this design on a frozen video DiT, with pseudo radial-distance supervision from a monocular geometry foundation model stabilizing the ray-position channel and probe-selected middle-layer placement keeping the adapter compact. In experiments, CRePE improves camera control over ReCamMaster and UCPE across pinhole and non-pinhole lenses with competitive video quality, and outperforms a RayRoPE-style endpoint approximation in controlled positional-encoding ablations, isolating the benefit of projected-path integration over endpoint-based approximations.

Beyond camera-only conditioning, the same positional-encoding pathway, trained with Radial MixForcing, exposes a controllable interface for external scene geometry. This unifies camera and scene-geometry control under a single adapter and enables scene-geometry-conditioned generation with quantitatively verified map adherence and source-video motion transfer with new prompts and target trajectories. We hope CRePE serves as a useful building block for geometry-aware, camera-controllable video generation under diverse lens configurations.

\begin{ack}
This work was supported by the National Research Foundation of Korea under Grant \mbox{RS-2024-00336454}, the Advanced GPU Utilization Support Program funded by the Government of the Republic of Korea (Ministry of Science and ICT) (\mbox{02-26-01-0404}), the AI Computing Infrastructure Enhancement (GPU Rental Support) User Support Program funded by the Ministry of Science and ICT (MSIT) Republic of Korea (\mbox{RQT-25-120217}), and Artificial intelligence industrial convergence cluster development project funded by the Ministry of Science and ICT (MSIT, Korea) \& Gwangju Metropolitan City.
\end{ack}

{\small
\bibliographystyle{unsrtnat}
\bibliography{references}
}

\newpage
\appendix

\section{Standard RoPE and UCM Details}
\label{app:rope_ucm}

Rotary positional encoding (RoPE)~\cite{su2023roformerenhancedtransformerrotary} encodes a scalar position $x$ as a block-diagonal rotation
\begin{equation}
    \rho_D(x) = \bigoplus_{f=1}^{D/2}\rho_2(\omega_f x),\qquad
    \rho_2(\omega x)=
    \begin{bmatrix}
    \cos(\omega x)&-\sin(\omega x)\\
    \sin(\omega x)&\phantom{-}\cos(\omega x)
    \end{bmatrix}.
\end{equation}
For multi-dimensional coordinates, we allocate separate channel groups to each coordinate.

The Unified Camera Model uses distortion parameter $\xi$ and projection
\begin{equation}
    \Pi^{\mathrm{UCM}}_\xi(\mathbf{X})=
    \left(f_x\frac{X}{Z+\xi\|\mathbf{X}\|}+c_x,\;
          f_y\frac{Y}{Z+\xi\|\mathbf{X}\|}+c_y\right),
\end{equation}
and inverse ray map
\begin{equation}
    \Phi^{\mathrm{UCM}}_\xi(x,y)=\mathrm{normalize}\left([\gamma x,\gamma y,\gamma-\xi]^\top\right),\quad
    \gamma=\frac{\xi+\sqrt{1+(1-\xi^2)(x^2+y^2)}}{1+x^2+y^2}.
\end{equation}
For $\xi\neq0$, the projection of a source-ray interval into a query view is generally curved, motivating CRePE's piecewise phasor integration.

\section{CRePE Implementation Details}
\label{app:crepe_details}

\begin{figure}[!htbp]
    \centering
    \includegraphics[width=0.92\linewidth]{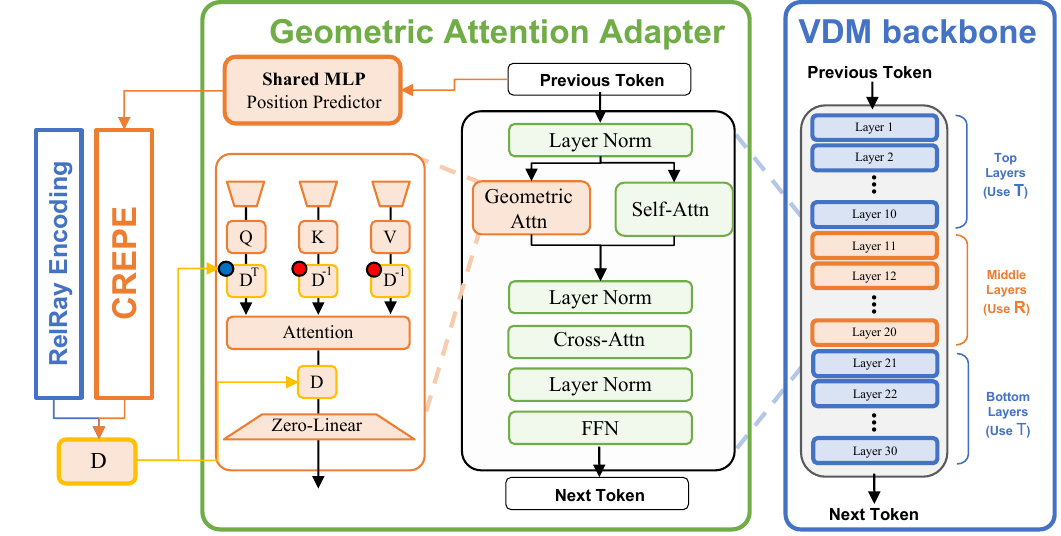}
    \caption{\textbf{Architecture of the Geometric Attention Adapter.}
    Our adapter follows the UCPE-style camera-conditioning structure, but augments the attention
    pathway with a CRePE-based geometric attention branch. The ``Shared MLP Position Predictor'' is the geometry head $g_\phi$. Following the layer-placement ablation,
    CRePE is applied to the middle 10 transformer blocks, where recoverable radial-distance
    information is strongest. The remaining early and late blocks retain the RelRay encoding used
    in UCPE. This hybrid design preserves the stable ray-only camera signal in the outer blocks
    while injecting depth-aware curved-ray positional modulation into the middle blocks.}
    \label{fig:supp_architecture}
\end{figure}

\paragraph{Backbone and training.}
Unless otherwise specified, the following settings describe the default 10K-step model.
The backbone is Wan2.1-T2V-1.3B~\cite{wan2025wan} with 30 transformer blocks, kept frozen.
We train the adapter on 81-frame clips at $480\times832$ resolution with global batch size 8,
AdamW, learning rate $10^{-4}$, gradient checkpointing, and 10K optimization steps.
All experiments run on a single NVIDIA B200 GPU; a default 10K-step training run takes approximately 3--7 days of wall-clock time, and the reduced-budget ablations (1K steps) use the same hardware.
CRePE is applied to blocks 11--20, numbering blocks 1--30 as in Figure~\ref{fig:supp_architecture}; other blocks retain UCPE-style relative-ray modulation.
The attention compression factor is 8, following~\cite{zhang2025unified}.
Figure~\ref{fig:supp_architecture} illustrates the overall Geometric Attention Adapter architecture.

\paragraph{Query--source frame pairs.}
Training clips are sampled purely by temporal position, not by relative camera motion: each example is a fixed-length clip of contiguous frames at the native frame rate. Attention operates densely over all query--source frame pairs, so every temporal offset up to the full clip span is present. Relative motion therefore ranges from near-degenerate adjacent pairs (median rotation $0.91^\circ$) to full-span pairs with substantial displacement (median $16.4^\circ$), the regime in which projected-path integration is most active.

\paragraph{Geometry head.}
The shared geometry head $g_\phi$ is implemented as an MLP, LayerNorm $\to$ Linear $\to$ SiLU $\to$ Linear, with hidden width $\max(16,D/4)$.
Radial-distance supervision is applied to its predictions at a single selected layer, training the same geometry head used across all CRePE-modulated layers.
The final layer is initialized with zero weights and bias $[0,3]$, giving $\mu=0$ and a broad initial interval $\sigma=3$.
The predicted log radial-distance interval is bounded to $[-3,3]$, corresponding to normalized radial distances in approximately $[0.05,20]$.
This bound limits the predicted CRePE interval on the normalized scale.
It is separate from pseudo-label validity filtering: raw metric pseudo radial-distance targets above $20\,\mathrm{m}$ are marked invalid rather than clipped to the upper bound.

\paragraph{Multi-ray patch encoding.}
Each token uses $A=3$ offset rays at the patch corners, with offsets $(0,0)$, $(1,0)$, and $(0,1)$ in patch units.
For offset $a$, the source ray is $\bm{r}_{s,p,a}=\Phi^{\mathrm{UCM}}_{\xi_s}(u_{p,a},v_{p,a})$.
We compute CRePE coefficients independently for the offsets and allocate disjoint RoPE channel groups to them.

\paragraph{Query-view coordinate.}
Given lifted point $\mathbf{X}^{(k)}_q=(X,Y,Z)^\top$ in the query camera, let $\beta_q=Z+\xi_q\|\mathbf{X}^{(k)}_q\|$.
We define
\begin{equation}
    \bar{u}^{(k)}=\frac{f_x}{W}\frac{X}{\beta_q},\qquad
    \bar{v}^{(k)}=\frac{f_y}{H}\frac{Y}{\beta_q},
\end{equation}
and use the bounded projected coordinate
\begin{equation}
    \mathbf{u}^{(k)}=
    \left(\frac{[\bar{u}^{(k)},\bar{v}^{(k)},1]^\top}{\|[\bar{u}^{(k)},\bar{v}^{(k)},1]\|}\right)_{1:2}.
\end{equation}
The final CRePE coordinate is $\tilde{\mathbf{x}}^{(k)}=(\mathbf{u}^{(k)},\|\mathbf{X}^{(k)}_q\|)$.
In addition, as in RayRoPE, each key token encodes the source camera center expressed in the query camera frame (3 coordinates); this term does not depend on the radial distribution and uses a single rotary phase, and for query tokens it is the origin. A token therefore carries $3+A\times(2+1)=12$ coordinates. Each coordinate uses 8 log-spaced RoPE frequencies with periods from $P$ down to $P/3^{7}$, where $P=8$ for the bounded image coordinates, $P=80$ for the radial range (clamped to $[-10,10]$), and $P=4$ for the camera center, so the 12 coordinates fill the 192-dimensional geometric-attention head.

\paragraph{Numerical stability.}
When $\theta^{(k+1)}-\theta^{(k)}$ is small in Eq.~\eqref{eq:phasor_main}, we use the limiting value of the segment integral.
Invalid projected points and masked pseudo labels are excluded from the auxiliary radial-distance loss.

\section{Layer Placement Ablation}
\label{app:layer_ablation}

\paragraph{Probe protocol.}
For each frozen Wan2.1 block, we train an independent lightweight probe with the same input features used by the adapter to predict the valid, normalized UniK3D radial-distance targets on the token grid. The backbone and video diffusion objective are not updated during probing; the probe is optimized only with the radial-distance supervision loss and evaluated on held-out PanShot clips. Lower probing error indicates that the layer exposes recoverable ray-position information to a small adapter head, not that the final generator has been trained with CRePE at that layer.

\paragraph{Placement selection.}
The probe curve is used as a selection guide rather than as the sole criterion. We evaluate contiguous CRePE windows on the first 10, middle 10, last 10, and all 30 DiT blocks under the reduced-budget protocol of the breakpoint ablation (1K steps, $\lambda_{\mathrm{rad}}=0$, pose stride 4, 50 clips). Middle-layer placement provides the best overall rank and the strongest pose/lens/orientation tradeoff, while retaining UCPE-style relative-ray modulation in the remaining blocks. This is why the default model applies CRePE to blocks 11--20 and uses UCPE modulation elsewhere.

\begin{figure}[!htbp]
\centering
\includegraphics[width=\linewidth]{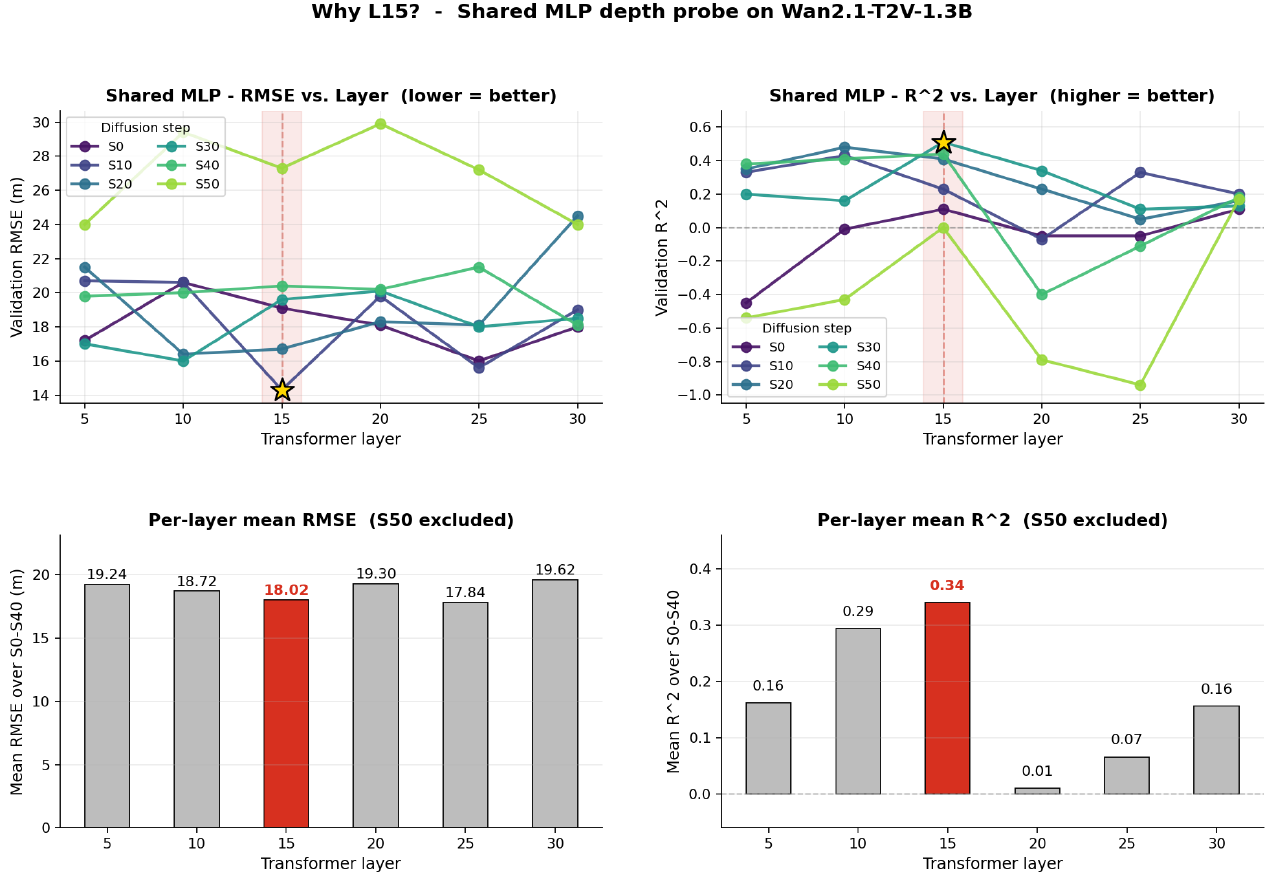}
\caption{\textbf{Layer-wise probing of radial-distance information.} Held-out radial-distance probing error per Wan2.1 layer. The shaded middle-layer window is used for CRePE.}
\label{fig:app_probing}
\end{figure}

\begin{figure}[!htbp]
\centering
\includegraphics[width=\linewidth]{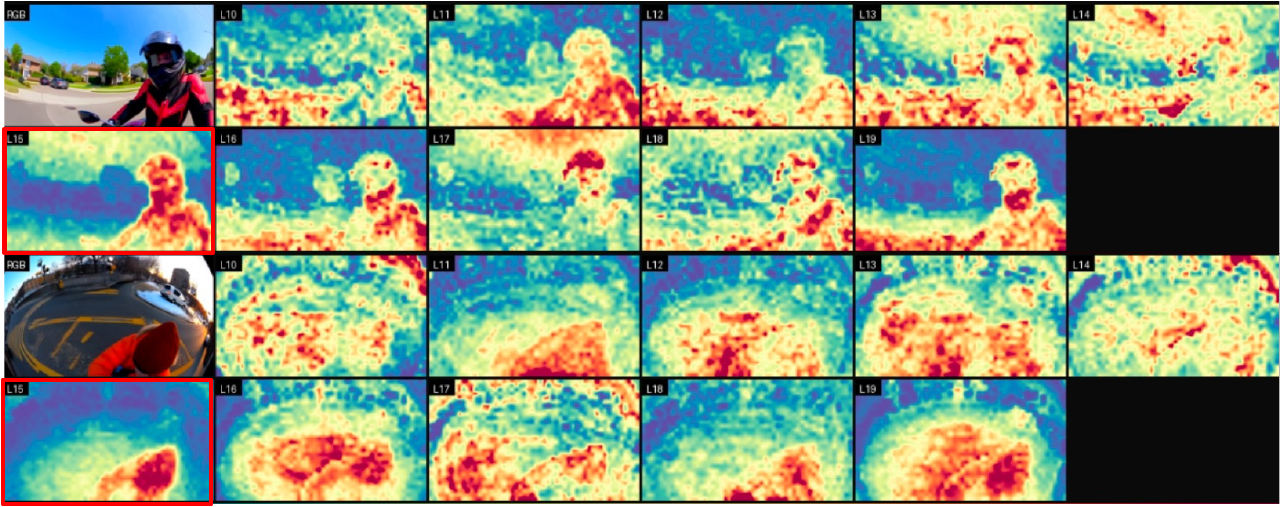}
\caption{\textbf{Layer-wise visualization of predicted radial-distance maps.}
We visualize radial-distance maps decoded from different Wan2.1 layers using the shared geometry head.
The highlighted map, from the layer used as the target-supervised layer in CRePE, shows the sharpest and most coherent scene structure.
This supports our choice of applying radial-distance supervision to the middle-layer representation selected by the probing analysis.
The maps visualize the token-resolution expected radial distance $\exp(\mu)$ that modulates attention; they are not depth-decoder outputs and should not be read as metric depth estimates.}
\label{fig:app_layerwise_depth}
\end{figure}

\begin{figure}[!htbp]
    \centering
    \includegraphics[width=\linewidth]{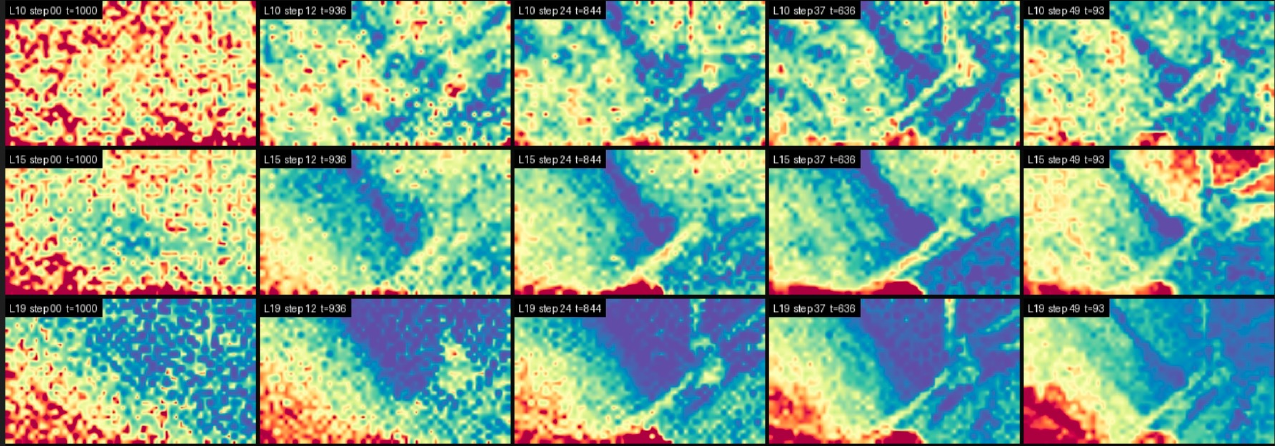}
    \caption{\textbf{Evolution of predicted radial-distance maps during denoising.}
    We visualize CRePE's internal radial-distance maps at multiple denoising steps.
    Denoising progresses from left to right. The maps evolve from coarse responses to
    structured scene layouts, suggesting that the ray-position pathway progressively
    organizes scene geometry during generation. As in Figure~\ref{fig:app_layerwise_depth},
    these are token-level latent radial distributions, not metric depth predictions.}
    \label{fig:radial_distance_denoising}
\end{figure}

Figure~\ref{fig:app_probing} reports the layer-wise radial-distance probing results. Figure~\ref{fig:app_layerwise_depth} visualizes decoded radial-distance maps across layers, Figure~\ref{fig:radial_distance_denoising} shows their evolution during denoising, and Table~\ref{tab:app_layer_placement} reports the layer-placement ablation.

\begin{table*}[!htbp]
\centering
\small
\setlength{\tabcolsep}{4pt}
\caption{\textbf{CRePE layer placement} (reduced-budget protocol: 1K steps, $\lambda_{\mathrm{rad}}=0$, pose stride 4, 50-clip ablation subset; the Middle10 row is the $K=5$ run of the breakpoint ablation). Applying CRePE to the middle 10 blocks provides the best overall rank and the strongest pose/lens/orientation tradeoff.}
\label{tab:app_layer_placement}
\adjustbox{max width=\linewidth}{%
\begin{tabular}{lrrrrrrrrrrr}
\toprule
\multirow{2}{*}{Setting}
& \multirow{2}{*}{Avg. Rank}
& \multicolumn{1}{c}{Pose}
& \multicolumn{2}{c}{Text/Image}
& \multicolumn{2}{c}{Video}
& \multicolumn{2}{c}{Lens}
& \multicolumn{2}{c}{Orientation} \\
\cmidrule(lr){3-3}\cmidrule(lr){4-5}\cmidrule(lr){6-7}\cmidrule(lr){8-9}\cmidrule(lr){10-11}
& & Rot. Err $\downarrow$ & CS-I $\uparrow$ & CS-T $\uparrow$ & FID $\downarrow$ & FVD $\downarrow$ & $k_1$ $\downarrow$ & $k_2$ $\downarrow$ & Pitch $\downarrow$ & Up $\downarrow$ \\
\midrule
All      & 3.06 & 4.877 & 99.27 & 25.70 & 136.8 & 1457 & 0.1610 & 0.09002 & 6.874 & 6.918 \\
First10  & 2.12 & 5.142 & 99.16 & \textbf{25.78} & 133.9 & \textbf{1438} & 0.1679 & 0.1113 & 6.732 & 6.902 \\
Middle10 & \textbf{2.00} & \textbf{4.649} & \textbf{99.30} & 25.76 & 136.6 & 1491 & \textbf{0.1417} & \textbf{0.08034} & \textbf{6.612} & \textbf{6.712} \\
Last10   & 2.82 & 4.843 & 99.21 & 25.68 & \textbf{130.3} & 1506 & 0.1574 & 0.08684 & 6.982 & 6.921 \\
\bottomrule
\end{tabular}}
\end{table*}

\section{Overall Camera-Generation Metrics}
\label{app:overall_metrics}

Table~\ref{tab:overall_full_comparison} reports the merged pinhole and non-pinhole results across pose, lens, orientation, text/image, video-quality, and VBench metrics.

\begin{table*}[!htbp]
\centering
\small
\setlength{\tabcolsep}{4.5pt}
\caption{\textbf{Overall camera-generation comparison.}
We merge pinhole and non-pinhole subsets and report overall metrics for ReCamMaster, UCPE, and CRePE.
Panel A reports camera-control, lens-distortion, and orientation metrics; Panel B reports text/image alignment and video-quality metrics; Panel C reports VBench quality and consistency metrics.}
\label{tab:overall_full_comparison}

\vspace{2pt}
\textbf{Panel A: Pose, lens distortion, and orientation}
\vspace{2pt}

\adjustbox{max width=\linewidth}{%
\begin{tabular}{lrrrrrrrrrr}
\toprule
\multirow{2}{*}{Method}
& \multicolumn{3}{c}{Pose}
& \multicolumn{2}{c}{Lens}
& \multicolumn{5}{c}{Orientation} \\
\cmidrule(lr){2-4}
\cmidrule(lr){5-6}
\cmidrule(lr){7-11}
& CamMC $\downarrow$
& Rot. $\downarrow$
& Trans. $\downarrow$
& $k_1$ $\downarrow$
& $k_2$ $\downarrow$
& Pitch $\downarrow$
& Roll $\downarrow$
& Gravity $\downarrow$
& Lat. $\downarrow$
& Up $\downarrow$ \\
\midrule
ReCamMaster
& 55.42
& 16.99
& 45.98
& 0.184
& 0.153
& 8.08
& 6.30
& 11.13
& 8.58
& 7.40 \\

UCPE
& 21.87
& 6.66
& 17.54
& 0.181
& 0.153
& 4.84
& 4.47
& 7.09
& 6.13
& 6.09 \\

\textbf{CRePE}
& \textbf{18.15}
& \textbf{6.33}
& \textbf{13.78}
& \textbf{0.156}
& \textbf{0.115}
& \textbf{4.57}
& \textbf{3.61}
& \textbf{6.24}
& \textbf{6.02}
& \textbf{5.88} \\
\bottomrule
\end{tabular}%
}

\vspace{6pt}
\textbf{Panel B: Text/image alignment and video quality}
\vspace{2pt}

\adjustbox{max width=\linewidth}{%
\begin{tabular}{lrrrrrrr}
\toprule
\multirow{2}{*}{Method}
& \multicolumn{2}{c}{Text/Image}
& \multicolumn{5}{c}{Video} \\
\cmidrule(lr){2-3}
\cmidrule(lr){4-8}
& CS-I $\uparrow$
& CS-T $\uparrow$
& FID $\downarrow$
& FVD $\downarrow$
& FVD-C $\downarrow$
& IS $\uparrow$
& IS Std. \\
\midrule
ReCamMaster
& 98.680
& 24.855
& 61.15
& 551.75
& 504.73
& 10.04
& 0.288 \\

UCPE
& 98.745
& 25.060
& 58.71
& \textbf{463.47}
& \textbf{440.40}
& \textbf{11.48}
& 0.230 \\

\textbf{CRePE}
& \textbf{98.772}
& \textbf{25.096}
& \textbf{58.08}
& 465.33
& 444.92
& 11.43
& 0.237 \\
\bottomrule
\end{tabular}%
}

\vspace{6pt}
\textbf{Panel C: VBench metrics}
\vspace{2pt}

\adjustbox{max width=\linewidth}{%
\begin{tabular}{lrrrrrr}
\toprule
\multirow{2}{*}{Method}
& \multicolumn{5}{c}{VBench} \\
\cmidrule(lr){2-6}
& Aesthetic $\uparrow$
& BG Cons. $\uparrow$
& Img. Qual. $\uparrow$
& Overall Cons. $\uparrow$
& Subject Cons. $\uparrow$ \\
\midrule
ReCamMaster
& 0.5599
& 0.9473
& 74.79
& 0.2142
& \textbf{0.9467} \\

UCPE
& 0.5617
& 0.9516
& 74.89
& 0.2157
& 0.9427 \\

\textbf{CRePE}
& \textbf{0.5634}
& \textbf{0.9524}
& \textbf{75.48}
& \textbf{0.2163}
& 0.9425 \\
\bottomrule
\end{tabular}%
}
\end{table*}

\section{Radial-Distance Supervision Details}
\label{app:radial_supervision}

Pseudo radial-distance maps are generated by UniK3D~\cite{piccinelli2025unik3duniversalcameramonocular}. We apply validity filtering before normalization. A raw metric pseudo target $\hat{r}^{\mathrm{m}}$ is invalid if UniK3D fails, is marked unreliable by our validity filters, is non-finite or non-positive, falls outside the supported distance range, or exceeds $20\,\mathrm{m}$. These far-field targets are not clipped to $20\,\mathrm{m}$; they are excluded from the loss and, under Radial MixForcing, fall back to the model prediction for that ray. We divide only valid metric maps by a per-clip near-distance statistic to obtain normalized targets, pool them to the token grid, and mask invalid regions. From $(\mu,\sigma)$, the normalized prediction is $\bar{r}=\exp(\mu)$. The uncertainty scale $s$ is derived from the variance of a uniform distribution over the normalized interval $[\exp(\mu-|\sigma|),\exp(\mu+|\sigma|)]$; this interval-width surrogate is separate from the log-uniform measure used for the phasor integration:
\begin{equation}
    v_{\mathrm{int}}=\frac{(\exp(\mu+|\sigma|)-\exp(\mu-|\sigma|))^2}{12}.
\end{equation}
We use $s=\sqrt{\max(v_{\mathrm{int}},10^{-6})}$ with a ceiling of 10, $\alpha=1.0$, and $\lambda_{\mathrm{rad}}=10^{-3}$. Supervision is gated off for the noisiest 3\% of diffusion timesteps.

\section{Additional Radial MixForcing Details}
\label{app:radial_mixforcing}

Section~\ref{sec:method:radial_mixforcing} gives the main Radial MixForcing algorithm. Here we keep only the implementation choices that are too detailed for the main text.

\paragraph{Teacher interval width.}
When a valid pseudo radial target is substituted, we use $\sigma_T=0.1$. Invalid pseudo targets never create a teacher interval: if the pseudo distance is unavailable, marked unreliable by the validity filters, non-finite or non-positive, outside the supported range, or above $20\,\mathrm{m}$ in raw metric units, the effective CRePE geometry for that ray remains $(\mu_{\mathrm{pred}},\sigma_{\mathrm{pred}})$.

\paragraph{Schedule.}
We use a piecewise-linear schedule that starts at full substitution probability, decays between steps 1000 and 7000, and then holds a small floor. For block-frame sampling, the floor is 0.1; for video-level sampling, the floor is 0.5. The mask is sampled once per forward pass and shared across all CRePE layers.

\paragraph{External-radial-map inference.}
For qualitative scene-geometry conditioning and source-video motion transfer, we intentionally provide an external radial map at inference and route it through the same CRePE pathway. The same validity and scale-normalization rule is used as in training: valid external radial values can overwrite the predicted radial tensor after normalization, while invalid or missing values use the model prediction for the corresponding ray. This changes only the positional coefficients, not the video backbone. For the source-video motion-transfer examples in Figure~\ref{fig:external_radial} (bottom), denoising starts from the source video encoded by the VAE and noised to strength $0.75$ (SDEdit-style~\cite{meng2022sdedit}, 50 steps) instead of pure noise, so part of the source appearance carries over in addition to the supplied geometry; the DAVIS evaluation in Table~\ref{tab:mix_quant} starts from pure noise.

\section{Dataset and Evaluation Details}
\label{app:dataset_details}

PanShot~\cite{zhang2025unified} contains panning videos with camera trajectories and UCM lens parameters. The train/test split prevents scene overlap. The dataset spans pinhole, wide-angle, and fisheye settings with horizontal FoV and UCM distortion. Camera controllability is measured by recovering camera motion from generated videos and comparing it to the requested trajectory. The main camera-generation comparison uses all 340 clips in the PanShot test split; we do \emph{not} remove the high-motion tail of the distribution (e.g., the top 20\% by camera-motion magnitude~\cite{zhang2025unified}), since large-trajectory clips are the regime in which curved-ray modeling is expected to matter most. Reduced-budget ablations use a fixed 50-clip subset of the test split (13 pinhole and 37 non-pinhole clips); the external-control subsets are described in Appendix~\ref{app:zeroshot_external}.

\section{Number of Breakpoints}
\label{app:k_ablation}

Table~\ref{tab:crepe_full_point_ablation} retains the expanded metric suite for $K\in\{2,5,7\}$. These variants follow the 1K-step, diffusion-only, stride-4 protocol in Appendix~\ref{app:curved_path_protocols}; they are distinct from the default 10K-step supervised model. Table~\ref{tab:ablation_num_points_with_pose} additionally includes $K=1$ and $K=3$ and reports ranks among all five settings.

\begin{table*}[!htbp]
\centering
\small
\setlength{\tabcolsep}{4.5pt}
\caption{\textbf{Full sampling-point ablation for CRePE.}
We report all available non-zero metrics for CRePE with different numbers of radial-distance breakpoints ($K=2$ is the endpoint approximation).
Panels A1--A3 summarize pose, text/image alignment, and video-quality metrics, while Panel B reports lens and orientation metrics.
Avg. Rank is computed over all displayed metrics except IS Std., among these three settings (Table~\ref{tab:ablation_num_points_with_pose} ranks five settings).
CRePE with $K=5$ achieves the best overall balance.}
\label{tab:crepe_full_point_ablation}

\vspace{2pt}
\textbf{Panel A1: Average rank and pose}
\vspace{2pt}

\adjustbox{max width=\linewidth}{%
\begin{tabular}{lrrrr}
\toprule
Setting & Avg. Rank $\downarrow$ & CamMC $\downarrow$ & Rot. $\downarrow$ & Trans. $\downarrow$ \\
\midrule
CRePE, $K=2$ & 2.55 & 14.75 & 5.648 & 11.09 \\
\textbf{CRePE, $K=5$} & \textbf{1.35} & \textbf{13.45} & \textbf{4.649} & \textbf{10.55} \\
CRePE, $K=7$ & 2.10 & 14.23 & 4.702 & 11.21 \\
\bottomrule
\end{tabular}%
}

\vspace{6pt}
\textbf{Panel A2: Text/image alignment}
\vspace{2pt}

\adjustbox{max width=\linewidth}{%
\begin{tabular}{lrrrrr}
\toprule
Setting & CS-I $\uparrow$ & CS-T $\uparrow$ & LPIPS $\downarrow$ & PSNR $\uparrow$ & SSIM $\uparrow$ \\
\midrule
CRePE, $K=2$ & 99.300 & 25.750 & 0.720 & 8.848 & 0.262 \\
\textbf{CRePE, $K=5$} & \textbf{99.301} & \textbf{25.762} & \textbf{0.719} & \textbf{8.886} & \textbf{0.264} \\
CRePE, $K=7$ & 99.265 & 25.696 & 0.720 & 8.868 & 0.263 \\
\bottomrule
\end{tabular}%
}

\vspace{6pt}
\textbf{Panel A3: Video quality}
\vspace{2pt}

\adjustbox{max width=\linewidth}{%
\begin{tabular}{lrrrrr}
\toprule
Setting & FID $\downarrow$ & FVD $\downarrow$ & FVD-C $\downarrow$ & IS $\uparrow$ & IS Std. \\
\midrule
CRePE, $K=2$ & \textbf{135.34} & 1563 & 1589 & 9.55 & 0.351 \\
\textbf{CRePE, $K=5$} & 136.61 & \textbf{1491} & \textbf{1465} & \textbf{10.43} & 0.570 \\
CRePE, $K=7$ & 135.46 & 1550 & 1502 & 9.98 & 0.308 \\
\bottomrule
\end{tabular}%
}

\vspace{6pt}
\textbf{Panel B: Lens distortion and orientation}
\vspace{2pt}

\adjustbox{max width=\linewidth}{%
\begin{tabular}{lrrrrrrr}
\toprule
\multirow{2}{*}{Setting}
& \multicolumn{2}{c}{Lens}
& \multicolumn{5}{c}{Orientation} \\
\cmidrule(lr){2-3}
\cmidrule(lr){4-8}
& $k_1$ $\downarrow$
& $k_2$ $\downarrow$
& Pitch $\downarrow$
& Roll $\downarrow$
& Gravity $\downarrow$
& Lat. $\downarrow$
& Up $\downarrow$ \\
\midrule
CRePE, $K=2$
& \textbf{0.1212}
& 0.0789
& 7.333
& 5.818
& 10.458
& 8.684
& 6.949 \\

\textbf{CRePE, $K=5$}
& 0.1417
& 0.0803
& \textbf{6.612}
& 5.598
& \textbf{9.688}
& \textbf{8.455}
& \textbf{6.712} \\

CRePE, $K=7$
& 0.1396
& \textbf{0.0779}
& 6.756
& \textbf{5.462}
& 9.801
& 8.588
& 6.757 \\
\bottomrule
\end{tabular}%
}
\end{table*}

\section{Protocols and Stratified Results for the Curved-Path Analysis}
\label{app:curved_path_protocols}

\paragraph{Supervision-free breakpoint protocol.}
All breakpoint variants in the main-paper ablation ($K\in\{1,2,3,5,7\}$) are trained for an identical reduced budget of 1K steps with the diffusion objective only ($\lambda_{\mathrm{rad}}=0$), the same parameter count, and the same data. UniK3D is used neither to train these variants nor at inference; layer probing only selects the placement window. All variants predict the same two interval parameters $(\mu,\sigma)$; for $K\ge2$, interior breakpoints are deterministic evaluations of this same interval, not additional learned geometry. Pose metrics for these ablations use frame stride 4 and are therefore not comparable in absolute terms to the stride-1 numbers of the main comparison tables.

\paragraph{Camera-model stratification.}
To examine the lens-dependent component of the curved-path gain, we stratify the supervision-free evaluation into pinhole ($\xi=0$, 13 clips) and non-pinhole ($\xi\neq0$, 37 clips) subsets. Table~\ref{tab:app_stratification} reports the raw errors together with the improvement of each curvature-capable setting relative to the endpoint approximation ($K=2$) within each subset. At $K=3$, translation gains are nearly identical across the two subsets (13.2\% vs.\ 13.4\%), suggesting a shared, lens-independent benefit of non-affine integration; CamMC and rotation gains are larger and more persistent on non-pinhole clips, consistent with an additional contribution from UCM-induced transverse curvature. The gains saturate rather than grow monotonically with $K$.

\begin{table}[!htbp]
\centering
\small
\setlength{\tabcolsep}{5pt}
\caption{\textbf{Pinhole/non-pinhole stratification of the supervision-free breakpoint ablation.} Percentages are error changes relative to $K=2$ within each subset (negative is better); all metrics are lower-is-better.}
\label{tab:app_stratification}
\adjustbox{max width=\linewidth}{%
\begin{tabular}{llrrr}
\toprule
Metric & Camera model & $K{=}2$ & $K{=}3$ & $K{=}5$ \\
\midrule
CamMC    & Pinhole     & 14.7457 & 13.1519 {\scriptsize($-10.8\%$)} & 14.7137 {\scriptsize($-0.2\%$)} \\
         & Non-pinhole & 14.7555 & 12.0174 {\scriptsize($-18.6\%$)} & 13.0098 {\scriptsize($-11.8\%$)} \\
\midrule
RotErr   & Pinhole     &  5.2380 &  5.0362 {\scriptsize($-3.9\%$)}  &  4.9312 {\scriptsize($-5.9\%$)} \\
         & Non-pinhole &  5.7914 &  4.2134 {\scriptsize($-27.2\%$)} &  4.5505 {\scriptsize($-21.4\%$)} \\
\midrule
TransErr & Pinhole     & 11.3287 &  9.8322 {\scriptsize($-13.2\%$)} & 11.7906 {\scriptsize($+4.1\%$)} \\
         & Non-pinhole & 11.0002 &  9.5238 {\scriptsize($-13.4\%$)} & 10.1079 {\scriptsize($-8.1\%$)} \\
\bottomrule
\end{tabular}}
\end{table}

\paragraph{Why this is not a sample-count effect.}
For normalized log-radius $t\in[0,1]$ and phase samples $\theta_k=\theta(t_k)$, the implemented coefficient of Eq.~\eqref{eq:phasor_main}, written $\bar\rho_K$ for $K$ breakpoints, is the average of exact per-segment integrals,
\begin{equation}
\bar\rho_K=\frac{1}{K-1}\sum_{k=1}^{K-1}
\frac{e^{i\theta_{k+1}}-e^{i\theta_k}}{i(\theta_{k+1}-\theta_k)}.
\label{eq:affine_invariance}
\end{equation}
If $\theta(t)=at+b$ is affine, every segment has the same phase increment, so the sum telescopes to $(e^{i\theta(1)}-e^{i\theta(0)})/[i(\theta(1)-\theta(0))]$, independent of $K$; extra subdivision helps only when the deployed phase is non-affine. Exponentiating log radius, camera translation, range encoding, and bounded coordinates create a shared non-affinity even for pinhole cameras, and the UCM denominator adds lens-dependent bending. Consistently, in a stratified evaluation (13 pinhole and 37 non-pinhole clips; Table~\ref{tab:app_stratification}), $K=3$ reduces CamMC over $K=2$ by $10.8\%$ on pinhole and $18.6\%$ on non-pinhole clips with nearly identical translation gains ($13.2\%$ vs.\ $13.4\%$), and at $K=5$ gains persist on non-pinhole clips (CamMC $-11.8\%$) but are small or mixed on pinhole clips. This indicates a shared viewpoint/depth benefit plus a non-pinhole component, rather than a purely UCM-curvature effect.

\section{Supervision and Robustness Details}
\label{app:supervision_robustness}

\paragraph{Supervision-duration study.}
We train the $K=5$ setting for 1K and 10K steps, with and without supervision. SSI-L1 is the scale- and shift-invariant L1 error between the geometry head's prediction and depth re-estimated from its own generated video (lower means more self-consistent). Pose metrics use stride 1, as in the main comparison tables. The 1K-step runs are evaluated on the 50-clip ablation subset and the 10K-step runs on the full 340-clip test split, so we compare settings within a budget. Table~\ref{tab:app_sup2x2} shows that at the 1K ablation budget supervision provides no camera-control advantage (the unsupervised point estimates are slightly ahead), so the supervision-free protocol above is a fair isolation of the formulation. Under long optimization the two settings diverge: at 10K steps the supervised head is markedly more self-consistent (SSI-L1 0.087 vs.\ 0.133), the camera-control ordering reverses, and the unsupervised variant falls below even depth-free UCPE (CamMC 22.19 vs.\ 21.87). The lightweight anchor ($\lambda_{\mathrm{rad}}=10^{-3}$) therefore stabilizes the radial channel under long training rather than producing the formulation-level gain.

\begin{table}[!htbp]
\centering
\small
\setlength{\tabcolsep}{6pt}
\caption{\textbf{Role of pseudo radial-distance supervision.} The same $K=5$ configuration trained with and without $\mathcal{L}_{\mathrm{rad}}$ at two budgets (1K: 50-clip ablation subset; 10K: 340-clip test split; pose stride 1).}
\label{tab:app_sup2x2}
\adjustbox{max width=\linewidth}{%
\begin{tabular}{lrrrr}
\toprule
Metric & 1K no-sup & 1K sup & 10K no-sup & 10K sup \\
\midrule
CamMC $\downarrow$   & 52.02 & 56.27 & 22.19 & \textbf{18.15} \\
RotErr $\downarrow$  & 18.00 & 18.21 &  8.11 & \textbf{6.33} \\
Gravity $\downarrow$ &  9.69 & 10.04 &  7.57 & \textbf{6.24} \\
SSI-L1 $\downarrow$  & 0.129 & 0.117 & 0.133 & \textbf{0.087} \\
\bottomrule
\end{tabular}}

\end{table}

\paragraph{Alternate pseudo-depth teacher.}
Few monocular models provide metric radial prediction under UCM, which motivated our choice of UniK3D. Retraining CRePE with UniDepthV2 supervision~\cite{piccinelli2026unidepthv2} (1K steps, same $\lambda_{\mathrm{rad}}$; pinhole subset, as UniDepthV2 lacks UCM-compatible metric radial distance under strong distortion; stride-4 pose) reaches performance comparable to the UniK3D-supervised counterpart: CamMC 15.41 vs.\ 15.23, RotErr 5.72 vs.\ 5.89, CS-T 25.94 vs.\ 25.80. Because supervision has little effect at this budget (Table~\ref{tab:app_sup2x2}), this comparison suggests that the results are not specific to UniK3D rather than showing that the long-training anchor transfers across teachers.

\paragraph{Depth-Concat control.}
The Depth-Concat baseline receives the same pseudo radial signal under the same auxiliary loss and the full 10K-step protocol: a shared MLP predicts the pseudo radial distance, and the prediction is passed through an additional projection MLP and concatenated to the token features ($+6$M parameters) instead of entering the positional phase. Depth-Concat trains healthily and is comparable to CRePE in quality (FVD 473.0 vs.\ 465.3; CS-T 25.08 vs.\ 25.10), yet fails specifically on pose, below even depth-free UCPE (CamMC 23.21 vs.\ 21.87; CRePE 18.15). What matters is how geometry enters attention---as a phase varying per query with the relative camera transform, not as a query-agnostic feature.

\paragraph{Dynamic-region behavior.}
Frame-wise monocular pseudo-depth is least reliable on moving objects. On 50 PanShot clips, we evaluate the radial prediction of one middle-window layer at a single denoising step, using moving-object masks from Segment Any Motion~\cite{huang2025segmentmotion} pooled to the $30\times52$ token grid (a token is dynamic if at least $50\%$ of its pixels move, static if none; invalid tokens are excluded). Statistics use identical normalization and pooling, with 95\% clip-level bootstrap confidence intervals (2000 resamples; 34/50 clips retain sufficient valid tokens); the metrics measure agreement with the supervision signal, not ground-truth depth. Dynamic tokens disagree more with the pseudo target (log-MAE 0.294 [0.228, 0.342] vs.\ 0.208 [0.194, 0.236]; $\delta_1$-style agreement 0.381 [0.312, 0.492] vs.\ 0.533 [0.475, 0.560]) but receive $1.51\times$ wider predicted uncertainty (mean $|\sigma|$ 1.180 vs.\ 0.779), avoiding static-level confidence in unreliable labels.

\paragraph{Temporal stability of the radial channel.}
Table~\ref{tab:app_temporal_stability} reports the median absolute frame-to-frame change of per-frame log-radial statistics over shared valid tokens (median over frames, then clips; brackets are 95\% clip-level bootstrap confidence intervals). All four paired CRePE--UniK3D differences are significant, and the SLAM-consistent ViPE reference~\cite{huang2025vipe} anchors every row: ViPE $<$ CRePE $<$ UniK3D, so CRePE is ordered between the pose-consistent reference and its frame-wise teacher rather than merely over-smoothed. CRePE's scale changes also track ViPE better than UniK3D's do (median-statistic drift disagreement 0.020 vs.\ 0.034, favoring CRePE on 82\% of supported clips). The radial channel filters, rather than copies, teacher jitter.

\begin{table}[!htbp]
\centering
\small
\setlength{\tabcolsep}{5pt}
\caption{\textbf{Temporal stability of per-frame log-radial statistics.} Median absolute frame-to-frame change (lower is more stable); brackets are 95\% bootstrap confidence intervals.}
\label{tab:app_temporal_stability}
\adjustbox{max width=\linewidth}{%
\begin{tabular}{lccc}
\toprule
Per-frame statistic & CRePE & UniK3D & ViPE (anchor) \\
\midrule
Median & 0.0156 [0.0156, 0.0234] & 0.0375 [0.0254, 0.0503] & 0.0084 \\
Q1     & 0.0269 [0.0220, 0.0312] & 0.0398 [0.0295, 0.0550] & 0.0080 \\
Q3     & 0.0181 [0.0156, 0.0234] & 0.0305 [0.0267, 0.0391] & 0.0091 \\
IQR    & 0.0234 [0.0156, 0.0244] & 0.0306 [0.0248, 0.0417] & 0.0114 \\
\bottomrule
\end{tabular}}
\end{table}

\section{Zero-Shot and External-Control Evaluation Details}
\label{app:zeroshot_external}

\paragraph{Zero-shot protocol and full results.}
Both zero-shot evaluations use the PanShot-trained models without any additional training or tuning, with pose metrics at stride 4. KITTI-360~\cite{liao2023kitti360} provides real $180^\circ$ fisheye driving videos (32 scenes); RealCam-Vid~\cite{zheng2025realcamvid} provides 100 test-split scenes from RealEstate10K, DL3DV, and MiraData9K (34/33/33). Main Table~\ref{tab:app_zeroshot_full}C reports the metrics measured on both datasets. CRePE improves every pose metric over UCPE on both, and VBench overall consistency also favors CRePE on both. On KITTI-360, GeoCalib orientation errors measured on the generated videos likewise favor the CRePE variants over UCPE: roll 1.668, gravity 2.688, and up-vector 2.654 for CRePE versus 2.076, 3.060, and 3.016 for UCPE, with pitch on par (1.783 vs.\ 1.777), and the no-supervision variant reaching pitch 1.478 and gravity 2.500. Even the no-supervision variant outperforms UCPE on KITTI-360 in pose---indicating that the curved-path formulation itself transfers under strong lens distortion---while supervision provides the margin on the diverse pinhole content of RealCam-Vid.

\paragraph{External-control protocol.}
Radial-map adherence is evaluated on 33 DAVIS-2016 sequences~\cite{perazzi2016davis} (dynamic pinhole videos disjoint from training), selected by a valid-token-ratio threshold fixed before comparing methods. All methods receive the same source control maps; geometry is re-estimated from each generated video with the same fixed UniK3D model and compared at the $480\times480$ pixel level and the $30\times30$ token level after $16\times16$ average pooling. Scale Jitter is the median absolute frame-to-frame change of the fitted per-frame depth-scale factor. The non-pinhole control evaluation uses the 48 of 50 non-pinhole PanShot scenes with sufficient valid depth (a set separate from the 50-clip ablation subset), with the radial map as the supplied control.

\paragraph{Teacher-width sweep.}
In Radial MixForcing, the injected radial map defines the mean of the teacher's radial distribution and $\sigma_T$ its width. Training uses $\sigma_T=0.1$; Table~\ref{tab:app_sigma_sweep} sweeps $\sigma_T$ at inference on the MixForcing-trained model. Per-metric optima spread across the range. Camera-motion error varies mildly over a $20\times$ change (CamMC spread 3.9\%), while video-quality and lens-distortion metrics vary more. The range $0.05$--$0.3$ maintains comparable camera-motion control, and raising $\sigma_T$ at inference can partially relax the rigidity of tightly forced geometry. CS-T rises monotonically but only by $+0.9\%$, so the flexibility--alignment trade-off is minor.

\begin{table}[!htbp]
\centering
\small
\setlength{\tabcolsep}{7pt}
\caption{\textbf{Inference-time teacher-width sweep for Radial MixForcing.}}
\label{tab:app_sigma_sweep}
\adjustbox{max width=\linewidth}{%
\begin{tabular}{lrrrr}
\toprule
Metric & $\sigma_T{=}0.05$ & $\sigma_T{=}0.10$ & $\sigma_T{=}0.30$ & $\sigma_T{=}1.00$ \\
\midrule
CamMC $\downarrow$   & \textbf{20.39} & 21.02 & 21.18 & 21.14 \\
$k_1$ $\downarrow$   & 0.148 & 0.148 & \textbf{0.116} & 0.163 \\
$k_2$ $\downarrow$   & 0.110 & 0.101 & \textbf{0.070} & 0.104 \\
FVD $\downarrow$     & 1104 & \textbf{1062} & 1079 & 1394 \\
CS-T $\uparrow$      & 25.73 & 25.80 & 25.81 & \textbf{25.96} \\
\bottomrule
\end{tabular}}

\end{table}

\section{Broader Impacts}
\label{app:broader_impacts}

CRePE makes camera motion and lens geometry more controllable in video generation. Positive uses include content creation with physically plausible camera moves, synthetic data for perception systems that rely on wide-angle and fisheye sensors (e.g., driving and robotics), and more geometrically consistent video simulation. The same controllability could also make synthetic videos more convincing and thereby easier to misuse for misleading or deceptive content. CRePE is a lightweight adapter on a publicly released backbone and inherits that backbone's capabilities and limitations; deployments should follow the backbone's usage policy and use standard safeguards such as content filtering and provenance marking of generated media. CRePE's internal radial distances are token-level latent variables supervised by monocular pseudo labels and should not be used as metric depth in safety-critical applications.

\end{document}